%% file: width_topo_new.tex
\documentclass[review,onefignum,onetabnum]{siamonline250106}

\input{ex_shared}

\ifpdf
\hypersetup{
  pdftitle={
  Topology-Guaranteed Image Segmentation: Enforcing Connectivity, Genus, and Width Constraints
  },
  pdfauthor={W. Li, X.-C. Tai, and J. Liu}
}
\fi

\begin{document}

\maketitle

\begin{abstract}
Existing research highlights the crucial role of topological priors in image segmentation, particularly in preserving essential structures such as connectivity and genus. Accurately capturing these topological features often requires incorporating width-related information, including the thickness and length inherent to the image structures. However, traditional mathematical definitions of topological structures lack this dimensional width information, limiting methods like persistent homology from fully addressing practical segmentation needs.
To overcome this limitation, we propose a novel mathematical framework that explicitly integrates width information into the characterization of topological structures. This method leverages persistent homology, complemented by smoothing concepts from partial differential equations (PDEs), to modify local extrema of upper-level sets. This approach enables the resulting topological structures to inherently capture width properties.
We incorporate this enhanced topological description into variational image segmentation models. Using some proper loss functions, we are also able to design neural networks that can segment images with the required topological and width properties. Through variational constraints on the relevant topological energies, our approach successfully preserves essential topological invariants such as connectivity and genus counts, simultaneously ensuring that segmented structures retain critical width attributes, including line thickness and length.
Numerical experiments demonstrate the effectiveness of our method, showcasing its capability to maintain topological fidelity while explicitly embedding width characteristics into segmented image structures.\end{abstract}



\section{Introduction}

Image segmentation is a fundamental task in image processing, involving the partitioning of a digital image into non-overlapping, meaningful regions. Although standard segmentation methods are effective at identifying object boundaries, they often struggle to preserve crucial topological properties, such as the number of connected components and holes. Even small topological errors can have a significant impact on downstream tasks. As a result, maintaining topological accuracy in segmented images is essential and challenging.

Image segmentation methods can be broadly classified into model-driven and data-driven approaches. Traditional model-driven segmentation techniques typically utilize predefined prior information specified by the models. Based on low-level image features such as grayscale, spatial texture, geometry, and other characteristics, images are divided into several disjoint regions, ensuring feature similarity within the same region and differences across different regions. Among model-driven methods, the variational approach, which minimizes \textcolor{black}{an} energy function, is widely employed. Variational models can be categorized into edge-based models, such as the snake model \cite{snake1988} and the geodesic active contour (GAC) model \cite{gac1997}, and region-based models, such as the Potts model \cite{potts1952,potts2021}, the Mumford-Shah (MS) model \cite{ms1989}, the Chan-Vese (CV) model \cite{cv2001}, the Molecular-Beam-Epitaxy (MBE) level set model \cite{mbe}, the Modica-Mortola phase transition model \cite{mmpt} and the thresholded-Rudin-Osher-Fatemi (T-ROF) model \cite{trof,ms-trof}. In addition, total variation (TV) \cite{tv} is one of the most widely used regularization terms in variational models. Numerous advanced methods have been developed to improve it, including threshold dynamics (TD) \cite{LIU20112093}, shearlet \cite{shearlet}, p-Laplacians \cite{KANG20141136} regularization, and the boundary term with thickness \cite{thicktv}. One advantage of the variational model is its ability to incorporate constraint properties as regularization terms in the energy functional, thereby preserving these properties in the segmentation result. With the increasing volume of data and advancements in hardware, data-driven image segmentation methods had rapidly evolved. Convolutional neural networks (CNNs) \cite{unet,unet++,deeplabv3+}, transformer-based models \cite{segformer,sam}, \textcolor{black}{and diffusion-based models \cite{segdiff, GOU2025107744}} demonstrated significant strengths in feature extraction. 
Despite their success in achieving high pixel-level accuracy, data-driven models typically treat segmentation as a classification problem at the pixel scale. This perspective makes it difficult to enforce the global topological properties of segmented objects, such as connectivity or the correct number of holes.

To address this limitation, the Distance Regularized Level Set Evolution (DRLSE) model was introduced by \cite{drlse}. It incorporated a distance penalty term and an external energy term that guided the contour evolution toward an optimal shape. This formulation encouraged the level-set function to evolve into a form resembling a signed distance function, thereby encoding width information in the segmentation process.

In a related effort, \cite{ddt} proposed a neural network designed to predict the distance transform of an image. This predicted transform was then used to smooth the image skeleton, enabling the segmentation to incorporate width priors effectively.

In addition to incorporating width information, preserving topological structures remains a central challenge for meaningful segmentation. This has motivated substantial research efforts in recent years, aimed at enhancing the topological fidelity of segmentation outcomes.
Firstly, connectivity in segmentation results can be improved using methods that extract texture and fine structural details. Techniques such as snake convolution \cite{snakeconv} and fractal dimension analysis \cite{fractal} have proven effective in this regard. Moreover, the VGG network has demonstrated success in capturing elongated structures. In \cite{vggtopo}, features of its intermediate layers were incorporated into the loss function to enhance the representation of slender structures. Building on this, \cite{twgan} proposed a loss function that combines the width information extracted by the network with the intermediate features of VGG to further improve the segmentation of fine structures.
Secondly, to enhance connectivity, non-linear relationships between pixels have been leveraged. For instance, \cite{cldice} preserved the morphological skeleton similarity between the segmentation outputs and the ground truth. Connectivity matrices \cite{biconnet, conn} and the maximin affinity learning approach \cite{affinity} were introduced to model pixel relationships and inform the design of loss functions.
Thirdly, discrete Morse theory has been applied to extract the skeleton and topological features of images. In \cite{dmtloss}, a loss function was proposed to guide the network's attention toward structurally significant regions. Additionally, \cite{topologyaware} employed persistent homology to filter out noise and preserve essential topological features.

While these methods improve segmentation connectivity, they primarily provide soft constraints that encourage networks to focus on structural features. However, they do not offer strong guaranties of topological consistency in the segmentation outcomes.
To address the challenge of maintaining topological structures in segmentation, several methods have been developed that provide mathematical guaranties of topological consistency.
One approach, presented in \cite{deformation}, employs image registration to preserve topology by applying prior deformations, leveraging a fast diffeomorphic registration algorithm \cite{diffeomorphic}. 
\textcolor{black}{We would like to particularly highlight the recent work~\cite{Lambert2024}, 
which presents a versatile and adaptable mathematical framework for encoding geometrical and topological constraints 
into a deep neural network. This is achieved by combining segmentation and registration within a unified formulation, 
and subsequently optimizing the problem through alternating updates of the segmentation network parameters and the deformation field.}
Another technique incorporated the Beltrami coefficient as a regularization term in classical segmentation models \cite{beltrami}. Building on this, \cite{ZHANG2021218} introduced a model that combines Laplacian smoothing and conformality distortion to ensure that the transformation remains smooth and quasi-conformal.

Further advances include a 3D topology-preserving registration-based segmentation framework with hyperelastic regularization proposed in \cite{hyperelastic}, and a data-driven model by \cite{quasiconformal}, which integrates a ReLU-Jacobian regularizer with a Beltrami adjustment module. 
While these registration-based methods offer robust theoretical guaranties for topological preservation, they are highly sensitive to the choice of initial values and often struggle to accurately segment complex tubular structures.
The construction of topological energies relies on persistent homology (PH) \cite{topo_comp}, 
which ensures topological consistency by constraining Betti numbers. 
PH has become one of the most widely used tools in topological data analysis (TDA) 
due to its ability to extract meaningful topological information from data. 
Initially, \cite{topocrf} addressed regions with topological errors by introducing perturbations 
to the unary potential function. Subsequently, PH-based topological energies have been incorporated 
into loss functions~\cite{cloughtopoloss,hutopoloss}, pre-processing~\cite{Vandaele2020}, 
and post-processing~\cite{postph}. 
However, several challenges remain in image processing applications. 

\begin{figure}[!ht]
    \centering    
    \subcaptionbox*{Image}
    {\includegraphics[width=0.171\linewidth]{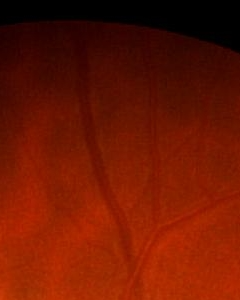}}
    \subcaptionbox*{Ground truth}
    {\includegraphics[width=0.25\linewidth]{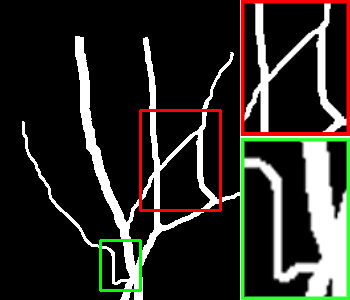}}
    \subcaptionbox*{UNet\cite{unet}}
    {\includegraphics[width=0.25\linewidth]{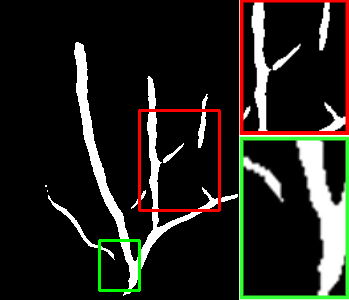}}
    \subcaptionbox*{PH\cite{ph}}
    {\includegraphics[width=0.247\linewidth]{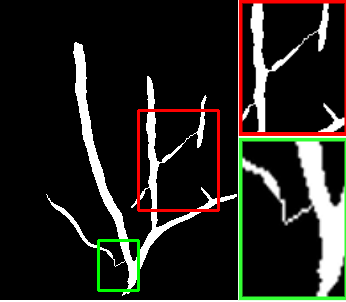}}
    \caption{Persistent homology energy to constrain one connected component. From left to right: image of CHASEDB1 \cite{chase}, ground-truth, results of UNet\cite{unet}, results of UNet with PH energy\cite{ph}. Compared to the result of UNet, PH links the unconnected regions by a single-pixel-width line.}
    \label{fig:cha}
\end{figure}

The PH approach transforms an image into a simplicial complex to compute its homology groups, 
preserving topological properties while neglecting geometric attributes such as width. 
Specifically, PH ensures topological consistency by constructing topological energies 
defined at critical points, which are constrained by individual pixels. 
When an image is represented as a \textcolor{black}{simplicial} complex, 
two connected subcomplexes have a nonempty intersection, implying that the intersection 
is at least a 0-dimensional simplex corresponding to a pixel. 
Consequently, a single-pixel line is sufficient for two regions in the image to be 
topologically connected, as illustrated in Figure~\ref{fig:cha}. 

Several other methods have encountered similar issues. 
The tree prior \cite{tree} enforced connectivity by designating parent nodes 
and requiring foreground pixels to form a minimum spanning tree with respect to those nodes. 
In \cite{topolevelset}, topology was preserved by selectively removing or adding points 
from the digital image without altering its overall topological structure. 
\textcolor{black}{The approach in \cite{topostd} integrated this techniques with the soft-threshold method to approximate object boundary length originally proposed in \cite{LIU20112093} and extended by  \cite{WANG2017657}.
} 
Similarly, \cite{homotopy} modified points via homotopy warping to guarantee that no topological changes occur. These methods are generally limited to binary images and may produce single-pixel artifacts that remain connected or create unintended holes.

However, these methods do not sufficiently capture the topological requirements of images, 
highlighting the need to incorporate width information while preserving topological consistency. 
\textcolor{black}{In medical imaging, single-pixel-width lines may achieve topological connectivity 
when connecting blood vessels, but in practice, such artificially thin vessels can compromise 
blood perfusion, thereby affecting physicians’ diagnosis, hemodynamic analysis, 
and other downstream tasks. Similarly, in remote sensing, accurate road width information 
is essential for navigation, route planning, and related applications.} 
To address this issue, \cite{topogan} proposed an insightful approach that integrates 
the distance transformation function with a topological loss constructed using 
persistent homology (PH). 
This method effectively constrains the topological consistency of the image after distance transformation; however, it may still introduce topological inaccuracies 
in the original image.

Since in image segmentation, the bandwidth information of the topological structure in the segmentation results is crucial. Given that none of the existing works have considered preserving the bandwidth information of the topological structure from a mathematical theory perspective, in this paper, we introduce a novel approach that employs the persistent homology method to ensure topological consistency while preventing critical points from appearing or disappearing in isolation. By ensuring that critical points are smooth within their neighborhoods, we simultaneously constrain both topological and width information. Therefore, we can integrate the width information of the image into the persistent homology method while genuinely preserving the topology.

Overall, our contributions can be summarized as follows:
\begin{itemize}
\item We introduce a novel topological energy based on persistent homology that incorporates width information, enabling the preservation of structural thickness in image segmentation. Traditional homology captures only topological invariants, while discarding geometric details such as width. To overcome this, we smooth the critical points of persistent homology using the previously introduced morphological gradient. This allows the birth and death of topological features to be influenced by their local neighborhoods, thereby embedding width information while maintaining topological consistency.

\item We propose a framework for integrating this width-aware topological energy into both variational model-based and data-driven deep neural network for image segmentation. Numerical experiments validate the effectiveness of our approach in preserving both topology and structural width.
\end{itemize}

The remainder of this paper is structured as follows. \Cref{sec:relatework} presents a review of related models and definitions. Subsequently, in \cref{sec:method}, we introduce our width-aware topological method, which is based on persistent homology method and smooth morphological gradients. Then, this section details the combination of the width-aware topological energy and segmentation models. The numerical results are presented to demonstrate the high validity of the proposed method in traditional and data-driven models in \cref{sec:experiment}. Finally, we draw conclusions and outlooks in \cref{sec:conclution}.

\section{The related works}\label{sec:relatework}

\subsection{Variational segmentation model}
The variational approach is one of the mainstream model-driven methods for image segmentation. In this method, segmentation tasks can be seen to divide the pixel domain $\Omega$ into many distinct parts. Therefore, its subdomain $\Omega_{\ell}$ can be expressed in terms of its relaxed  indicator function \textcolor{black}{$u_{\ell}: \Omega \rightarrow [0,1]$. In the region $\Omega_\ell$, the value of $u_\ell$ is close to 1, and otherwise $u_\ell$ is close to 0.} Then the segment condition can be written in the simplex form:
\begin{align}\label{eq:u_condition}
    \mathbb{U}=\left\{\boldsymbol{u}=\left(u_{1}, \ldots, u_{L}\right) \in[0,1]^{L}: \sum_{\ell=1}^{L} u_{\ell}(x)=1, \forall x \in \Omega \right\}. 
\end{align}
Furthermore, the Potts model \cite{potts1952} and its relaxed version \cite{potts2021} given below are one of the most fundamental segmentation models used in the literature:
\textcolor{black}{
\begin{equation} \label{eq:potts}
    \begin{aligned}
      \min_{\bm u \in \mathbb{U}} - \underbrace{\sum_{\ell=1}^{L} \int_{\Omega} o_{\ell}(x)u_{\ell}(x) \mathrm{d} x}_{\mathcal{F}}+\underbrace{\lambda \sum_{\ell=1}^{L}\left|\partial \Omega_{\ell}\right|}_{\mathcal{R}},
    \end{aligned}
\end{equation}
}
where $\lambda>0$ is a regularization parameter and $L$ is the total number of segmentation categories. In the Potts model, $\mathcal{F},\mathcal{R}$ represent the data fidelity term and the regularization term, respectively. 
This model is in fact a continuous min-cut problem which is equivalent to a continuous max-flow problem, see  \cite{potts2021} for a more detailed survey on this topic. 
The data fidelity term measures the degree of similarity between each class of segmentation results $u_\ell$ and image features $o_{\ell}$.

For the regularization term, the total variation (TV) \cite{tv} is the most widely applied. However, considering the non-smooth and high calculation cost of TV, the threshold dynamics (TD) regularization was proposed, which represents the boundary by the difference of values within the neighborhood. The idea of TD originated from spatially based Markov random fields \cite{tdmrf} and has since been applied as a spatial regular term to segmentation models \cite{fuzzy, cvt, LIU20112093, gmm, WANG2017657}. Its advantage is that it can be efficiently solved by the MBO splitting scheme \cite{mbo} and applied to the PDE-based image segmentation method \cite{tdms}. The TD energy to
 approximate the boundary length $|\partial\Omega_i|$ can be written as:
\begin{equation*}
|\partial\Omega_\ell|\approx\sqrt{\frac{\pi}{\sigma}}\sum_{\ell=1}^{L}\int_{\Omega}u_\ell(x) (k*(1-u_\ell))(x) \mathrm{d}x,
\end{equation*}
where $k$ is the indicator function of a ball of radius $\sigma$ as in \cite{LIU20112093} or  the Gaussian kernel $k(x)=\frac{1}{(2\pi\sigma^2)}e^{-\frac{|x|^2}{2\sigma^2}}$ parameterized by $\sigma$ as in \cite{WANG2017657} . The symbol $*$ represents the convolution operator. When $\sigma\rightarrow 0$, the TD regularization $\Gamma$-converges to $|\partial\Omega_\ell|$ \cite{tdconverge2007}.

Based on the above derivations, \cite{std} proposed the following Soft Threshold Dynamics (STD) model to approximate the relaxed Potts model (\ref{eq:potts}):
\begin{equation}\label{eq:std}
    \min\limits_{\bm u\in\mathbb{U}}\left\{\underbrace{\langle-\bm o,\bm u\rangle+\gamma\langle\bm u,\ln \bm u\rangle}_{:=\mathcal{F}(\bm u)}\right. 
    \left. +\underbrace{\lambda\langle \bm u, k*(1-\bm u)\rangle}_{:=\mathcal{R}(\bm u)}\right\}. 
\end{equation}
Here, $\mathcal{F}(\bm u)$ can be obtained from the log-sum-exp function after twice Legendre transformation, and $\mathcal{F}(\bm u)$ is a functional of the Softmax function in the variational problem . Throughout the article, $\langle \cdot,\cdot \rangle$ is the $L^2$ inner product for scalar/vector functions, that is, $\langle \bm o,\bm u \rangle =\sum_{\ell=1}^L \int_{\Omega} o_\ell(x)u_\ell(x) \mathrm{d}x.$

\subsection{Mathematical morphology for image processing}

Mathematical morphology is a method based on set theory which consists of the object and the structuring element (SE). As it analyzes and processes the structure of the target object by structuring element, it has a wide range of applications in the fields of feature extraction, edge detection, and others \cite{morph1987, morph1996, morph2006}. Let $\Omega$ and $\mathbb{B}$ represent the coordinate space of the image and the structuring element. Besides, considering a grayscale image $u(\bm x), \bm x \in \Omega$ and a flat structuring element $b(\bm z), \bm z \in \mathbb{B}$, where $\mathbb{B}$ is a neighborhood centered at the origin. Then the traditional morphological operators for grayscale images have the following definitions:

\begin{definition}[Grayscale erosion and dilation \cite{morphgradient}]\label{def:eros_dila}
    The erosion of an image $u$ by structuring element $b$ at position $\bm x$ is denoted:
    $$ [u \ominus b] (\bm x) = \min_{\bm{z} \in \mathbb{B}} \{ u(\bm{x} + \bm{z}) \},$$
    the dilation of the image $u$ by structuring element $b$ at position $\bm x$ is denoted:
    $$ [u \oplus b] (\bm x) = \max_{\bm{z} \in \mathbb{B}} \{ u(\bm{x} + \bm{z}) \}.$$
     Here, $\bm{x} + \bm{z} \in \Omega$, and $\Omega$ is the image domain.
\end{definition}

Definition \ref{def:eros_dila} means that the grayscale erosion of $u$ by the structuring element $b$ at $\bm x$ is the minimum value of the image in the neighborhood that coincides with $b$ and the dilation is the maximum value. Based on definition \ref{def:eros_dila}, we have the definition of internal and external gradients:

\begin{definition}[Internal gradient and external gradient \cite{morphgradient}]\label{def:mor_grad}
    For $\bm{x} \in \Omega$, the internal gradient $g^{-}$ is defined as the difference between the origin image and the eroded image:
    $$ [g^{-}(u)](\bm{x}) = u(\bm{x}) - [u \ominus b](\bm x),$$
    the external gradient $g^{+}$ is defined as the difference between the dilated image and the origin image:
    $$ [g^{+}(u)](\bm{x}) = [u \oplus b](\bm x) - u(\bm{x}).$$
\end{definition}

Similarly to the TV regular term, miniaturizing the outer and inner gradients in morphology can constrain the image piecewise smooth; therefore, we utilize the method to smooth the persistent homology critical points, which will not born or disappeared in isolation.
 
\subsection{Persistent homology}
\textcolor{black}{This section aims to provide an intuitive understanding of persistent homology. We present additional topological concepts and further details on persistent homology in appendix \ref{apdx:ph}. Moreover, readers unfamiliar with persistent homology may refer to some comprehensive introductions and surveys \cite{comp_topo, topo_comp, gudhi}.}

\textcolor{black}{Homology theory provides a rigorous algebraic framework for characterizing the topological structure of a space by quantifying the number of ``holes" across different dimensions through the computation of the dimensions of homology groups, known as Betti numbers. Subsequently, persistent homology extends classical homology theory to the analysis of real-world data. Its core innovation lies in the introduction of a scale parameter, which transforms static topology into a dynamic topological analysis. By tracking changes in this parameter, it assigns a lifetime to topological features, where each feature is associated with a birth value $b$ and a death value $d$ (a simple example is illustrated in \cref{fig:ph_ex}). We refer to each pair $(b, d)$ as a critical value pair.}

For image segmentation tasks, the vast majority of the work is based on pairs of critical values. \cite{hutopoloss} improved topological accuracy by constructing a loss function that minimizes the Wasserstein distance between the segmentation result and the ground truth. Besides, it could also improve the topological accuracy by post-processing \cite{cloughtopoloss} the segmentation result by minimizing the energy, \emph{etc.}. The idea of the persistent homology method is to keep the desired homology group and kill the undesired ones, while the homology group loses the original width information of the image and only retains the topological features, resulting in the image itself appearing distorted even if topological consistency is achieved in a topological sense. To improve this situation, we propose the following method.

\section{The proposed theoretical and numerical approaches} \label{sec:method}
The thickness of the topological structures in the image segmentation results reflects the width of the lines and boundaries. We use morphological gradients to improve the original persistent homology method.

\subsection{Smooth form of morphological gradient} \label{sec:smoothmorph}

Since the $\min$ and $\max$ functions contained in the definition \ref{def:eros_dila} are non-differentiable in mathematics, we propose the smooth forms of erosion and dilation. Smooth erosion and dilation operations correspond to smoother external gradient fields, as well as internal gradient fields, which are more stable for backpropagation in data-driven image segmentation. In addition, the outer and inner boundaries are binary forms of the external and internal morphological gradients. For grayscale images, the smooth morphological gradient provides better control over the width of the boundaries. The following is a detailed mathematical derivation process. 

For convenience in this section, we denote $\mathbb{B}(x,r)$ \textcolor{black}{as} a neighborhood centered on $x$ with radius $r > 0$, \textcolor{black}{that is, $\mathbb{B}(x,r) = \{y ~|~ d(x,y) \le r  \}$, where $d(x,y)$ denotes the distance between points $x$ and $y$.} 
%
In continuous form, as shown in appendix \ref{apdx:smoothmorph},  the  dilation can be written as:
 \begin{equation} \label{eq:logsumexp}
     \max_{y \in \mathbb{B}(x,r)} u(y) = \lim_{\varepsilon \to 0^{+}} \underbrace{\varepsilon \ln \int_{\mathbb{B}(x,r)} e^{\frac{u(y)}{\varepsilon}} \mathrm{d}y}_{\mathcal{M}_{\varepsilon}(u)}. 
 \end{equation}
 As $\mathcal{M}_{\varepsilon}(u)$ is convex with respect to $u$, we have $\mathcal{M}_{\varepsilon} = \mathcal{M}^{**}_{\varepsilon}$. Thus,
\begin{align}\label{eq:smooth_max}
    \mathcal{M}_{\varepsilon}(u) =\mathcal{M}^{**}_{\varepsilon}(u) = \max_{k \in \mathbb{K}} \{ \langle  k,u \rangle_{\mathbb{B}(x,r)} - \varepsilon \langle  k,\ln k \rangle_{\mathbb{B}(x,r)} \},
\end{align}
where
\begin{equation*}
    \left\langle  k, u \right\rangle_{\mathbb{B}(x,r)} = \int_{\mathbb{B}(x,r)} k(y)u(y) \mathrm{d}y, 
\quad     \mbox{  and  } \quad 
\textcolor{black}{
    \mathbb{K}=\left\{k: \mathbb{B}(x,r)\to [0,1] \mid \int_{\mathbb{B}(x,r)} k(y)\mathrm{d}y = 1  \right\}.}
\end{equation*}
Similarly, \textcolor{black}{$\mathcal{M}_{\varepsilon}(-u)$ is convex, we have $\mathcal{M}_{\varepsilon}(-u) = \mathcal{M}^{**}_{\varepsilon}(-u)$. The smooth approximation of $\min$ is
\begin{align}
    \min_{y \in \mathbb{B}(x,r)}u(y)  &= -\max_{y \in \mathbb{B}(x,r)}(-u(y) ) = \lim_{\varepsilon \rightarrow 0^{+}}-\mathcal{M}_{\varepsilon}(-u) = \lim_{\varepsilon \rightarrow 0^{+}}-\mathcal{M}^{**}_{\varepsilon}(-u) \nonumber \\
    &=\lim_{\varepsilon \rightarrow 0^{+}}-\max_{k \in \mathbb{K}} \{ \langle  k,-u \rangle_{\mathbb{B}(x,r)} - \varepsilon \langle  k,\ln k \rangle_{\mathbb{B}(x,r)} \} \nonumber \\
    &=\lim_{\varepsilon \rightarrow 0^{+}}\min_{k \in \mathbb{K}} \{ \langle  k,u \rangle_{\mathbb{B}(x,r)} + \varepsilon \langle  k,\ln k \rangle_{\mathbb{B}(x,r)} \}.\label{eq:smooth_min}
\end{align}}
Furthermore, because \cref{eq:smooth_max} and \cref{eq:smooth_min} have a closed-form solution, we can define the smooth erosion and dilation kernel function that extends $k$ to $\Omega$. 
\begin{definition}[Smooth morphological kernel]\label{def:smooth_ker}
Given $u$, for all \textcolor{black}{$x\in\Omega$}, the kernel of smooth erosion is  
\begin{equation*}
    \textcolor{black}{k_{m}(x,y,\varepsilon)} = \left\{\begin{array}{cl}
Softmin(\frac{u(y)}{\varepsilon}), & y \in \mathbb{B}(x,r), \\
0, & y \in \Omega \setminus \mathbb{B}(x,r).
\end{array}\right.
\end{equation*}
The kernel of smooth dilation is  
\begin{equation*}
    \textcolor{black}{k_{M}(x,y,\varepsilon)} = \left\{\begin{array}{cl}
Softmax(\frac{u(y)}{\varepsilon}), & y \in \mathbb{B}(x,r), \\
0, & y \in \Omega \setminus \mathbb{B}(x,r).
\end{array}\right.
\end{equation*}
\color{black}{Here 
\begin{align*}
    Softmax(\frac{u(y)}{\varepsilon}) &= \frac{e^{\frac{u(y)}{\varepsilon}}}{\int_{\mathbb{B}(x,r)}e^{\frac{u(z)}{\varepsilon}}\mathrm{d} z}, \\
    Softmin(\frac{u(y)}{\varepsilon}) &= \frac{e^{\frac{-u(y)}{\varepsilon}}}{\int_{\mathbb{B}(x,r)}e^{\frac{-u(z)}{\varepsilon}}\mathrm{d} z}. 
\end{align*}}
\end{definition}

\textcolor{black}{For simplicity of notation, we use $k_{m}(x,y)$, $k_{M}(x,y)$ to denote $k_{m}(x,y,\varepsilon)$, $k_{M}(x,y,\varepsilon)$ in the following expressions. Next, the smooth external and internal gradients can be defined by the kernel functions above}:
\begin{definition}[Smooth external and internal gradient]
Let $u:\Omega\mapsto \mathbb{R}$, the smooth external gradient is 
  \begin{align}\label{eq:s_ex_grad}
    [g_{\varepsilon}^{+}(u)](x) = \int_{\Omega} \left(k_{M}(x,y)u(y) - \varepsilon k_{M}(x,y) \ln k_{M}(x,y)\right) \mathrm{d}y - u(x), 
  \end{align}
and the smooth internal gradient is 
  \begin{align}\label{eq:s_in_grad}
    [g_{\varepsilon}^{-}(u)](x) = u(x) - \int_{\Omega} \left(k_{m}(x,y)u(y) + \varepsilon k_{m}(x,y) \ln k_{m}(x,y)\right) \mathrm{d}y.
  \end{align}
\end{definition}

In fact, the first term of smooth dilation or erosion is the convolution of a spatially varying kernel function with $u$. The latter term ensures that the convolution kernel is smooth within a small neighborhood. Please note that $\varepsilon$ is a parameter to control the width of erosion or dilation. The smaller the value of $\varepsilon$, the narrower the width. In addition, the range of $\mathbb{B}(x,r)$ serves to control the width as well.

\subsection{Width-aware topological (WT) energy}
Since persistent homology is defined at critical points, when we use persistent homology to construct topological energy constraints on an image that has only one connected component, it usually connects a single-pixel line. This is clearly consistent with the notion of simply connected in the topological sense, but is not what is desired in image processing. Therefore, we constrain the critical points to have the smallest morphological gradient, and intuitively the critical points with topological changes will be piecewise smooth in the neighborhood of their structuring element. In this paper, it is worth noting that the symbols $b_j^k$ and $d_j^k$ represent the birth value and the death value, respectively, of the $j$-th topological feature in dimension $k$.  In addition, we consider the upper level set filtration $(b^k_j>d^k_j)$ and ignore those critical points where the critical value region of the death point is infinite. First, consider the most basic topological energy based on persistent homology:
\begin{equation}\label{eq:topo_B}
    \mathcal{B}(\bm b,\bm d,\beta_k)=\sum_{j=\beta_k+1}^{|\Im_k|} b^k_j-d^k_j,
\end{equation}
where $|\Im_k|$ denotes the maximum number of critical points of $k$-dimension persistence diagram. When $\mathcal{B}(\bm b, \bm d,\beta_k)$ reaches its minimization $0$, then the persistence lifetime of the topological structure with Betti number more than $\beta_k$ will be 0, and these topological structures are not allowed in this procedure. That is, the Betti number of $k$-dimension topological structure would not be exceeded $\beta_k$ when optimizing $\mathcal{B}(\bm b, \bm d,\beta_k)$ to its minimization. 

Furthermore, for the function $u_{\ell}: \Omega\rightarrow [0,1]$, let $\mathbb{X}_{b_j^k}=u_{\ell}^{-1}(b_j^k)$ be a birth critical points set of $k$-dimension topological structure, and $\mathbb{X}_{d_j^k}=u_{\ell}^{-1}(d_j^k)$ be a set of death critical points in which $\mathbb{X}_{b_j^k}$ has died. $b_j^k$ and $d_j^k$ are in one-to-one correspondence. Denote $u_{\ell}(y_j^k)=b_j^k$ and $u_{\ell}(z_j^k)=d_j^k$, so $(y_j^k,z_j^k)$ is the critical point of the critical value $(b_j^k,d_j^k)$. Thus, \cref{eq:topo_B} is equivalent to
\begin{align}
    \mathcal{B}(u_{\ell},\beta_k) &= \sum_{j=\beta_k+1}^{|\Im_k|} u_{\ell}(y_j^k)-u_{\ell}(z_j^k) \nonumber \\
    &= \int_{\mathbb{X}_{\bm b,\beta_k}} u_{\ell}(x)\mathrm{d}x - \int_{\mathbb{X}_{\bm d,\beta_k}} u_{\ell}(x)\mathrm{d}x \label{eq:topo_u}, 
\end{align}
where $\mathbb{X}_{\bm b,\beta_k}$ and $\mathbb{X}_{\bm d,\beta_k}$ are the coordinate space corresponding to the critical points of birth and death which $k$-dimensional Betti number greater than $\beta_k$, \textcolor{black}{that is, $\mathbb{X}_{\bm b, \beta_k} = \bigcup_{j=\beta_k+1}^{|\Im_k|} \mathbb{X}_{b_j^k}$ and $\mathbb{X}_{\bm d, \beta_k} = \bigcup_{j=\beta_k+1}^{|\Im_k|} \mathbb{X}_{d_j^k}$. Thus, on the one hand, we would like to encourage the persistence lifetime of the first $\beta_k$ topological structures, which can be achieved by maximizing energy $\mathcal{B}(u_{\ell},0)-\mathcal{B}(u_{\ell},\beta_k)$; on the other hand, beyond the first $\beta_k$ topological features that we encourage, we seek the vanishing of all others—namely those whose persistence time is shorter than that of the $\beta_k$-th feature, which is achieved through the minimization of energy $\mathcal{B}(u_{\ell},\beta_k)$.}

Combined with these two cases, it is possible to construct an energy term that constrains the Betti number to be exactly $\beta_k$:
\begin{align}
        \mathcal{T}(u_{\ell}):=& \color{black}{ \sum_{k=0}^1\mu_k\left(\mathcal{B}(u_{\ell},\beta_k) - [\mathcal{B}(u_{\ell},0) - \mathcal{B}(u_{\ell},\beta_k)]\right)} \nonumber\\
        =&\sum_{k=0}^1\mu_k\left(2\mathcal{B}(u_{\ell},\beta_k) - \mathcal{B}(u_{\ell},0)\right).\label{eq:phenergy}
\end{align}

\textcolor{black}{
Subsequently, we have utilized the smooth external gradient \cref{eq:s_ex_grad} and the smooth internal gradient \cref{eq:s_in_grad} to smooth the birth and death points in \cref{eq:topo_u}, respectively. The motivation is shown in \cref{rmk:moti}.
\begin{align*}
    \mathcal{B}_{\varepsilon}(u_{\ell},\beta_k) 
    &= \int_{\mathbb{X}_{\bm b,\beta_k}} u_{\ell}(x)\mathrm{d}x + \int_{\mathbb{X}_{\bm b,\beta_k}} [g_{\varepsilon}^{+}(u)](x) \mathrm{d}x \\
    &- \int_{\mathbb{X}_{\bm d,\beta_k}} u_{\ell}(x)\mathrm{d}x + \int_{\mathbb{X}_{\bm d,\beta_k}} [g_{\varepsilon}^{-}(u)](x) \mathrm{d}x \\
    &= \int_{\mathbb{X}_{\bm b,\beta_k}} \langle k_M,u_{\ell} \rangle - \varepsilon \langle k_M,\ln k_M \rangle \mathrm{d}x \\
    &- \int_{\mathbb{X}_{\bm d,\beta_k}} \langle k_m,u_{\ell} \rangle + \varepsilon \langle k_m,\ln k_m \rangle \mathrm{d}x.
\end{align*}
}Thus, we can obtain an energy that can force the Betti number to be equal to $\beta_k$ and retain the width information:
\begin{equation}\label{eq:wtenergy}
    \mathcal{T}_{\varepsilon}(u_{\ell}):=\sum_{k=0}^1\mu_k\left(2\mathcal{B}_{\varepsilon}(u_{\ell},\beta_k) - \mathcal{B}_{\varepsilon}(u_{\ell},0)\right).
\end{equation}
Here, $\mu_k>0$ is a parameter that controls the topological prior of dimension $k$. For example, $\mu_0=1,~\mu_1=0$ denotes that the connected component priors are activated, and $\mu_0=0,~\mu_1=1$ \textcolor{black}{indicates that a constraint is applied only to the genus, but not to the number of connected components.} To update the parameters to constrain the consistency of the topology through topological energies utilizing backpropagation methods, \cite{phdiff} proves that $\mathcal{B}(u_{\ell},\beta_k)$ is differentiable under a mild mathematical condition. \textcolor{black}{To facilitate computation, let $\mathbb{X}^{(t)}_{\bm b, \beta_k}$ and $\mathbb{X}^{(t)}_{\bm d, \beta_k}$ denote the birth and the death critical points, respectively, of the function $u_{\ell}^{(t)}: \Omega \to [0,1]$,} and
\textcolor{black}{
\begin{align*}
    \mathcal{Q}_{\varepsilon}(u_{\ell},u^{(t)}_{\ell},\beta_k) &= \int_{\mathbb{X}^{(t)}_{\bm b,\beta_k}} \langle k_M,u_{\ell} \rangle - \varepsilon \langle k_M,\ln k_M \rangle \mathrm{d}x \\
    &- \int_{\mathbb{X}^{(t)}_{\bm d,\beta_k}} \langle k_m,u_{\ell} \rangle + \varepsilon \langle k_m,\ln k_m \rangle \mathrm{d}x.
\end{align*}}
Then $\mathcal{Q}_{\varepsilon}(u^{(t)}_{\ell},u^{(t)}_{\ell},\beta_k) = \mathcal{B}_{\varepsilon}(u^{(t)}_{\ell},\beta_k)$. 

On these bases, we can calculate the variational form of $\mathcal{Q}_{\varepsilon}(u_{\ell},u^{(t)}_{\ell},\beta_k)$.
\begin{theorem}[Varitional of $\mathcal{Q}_{\varepsilon}(u_{\ell},u^{(t)}_{\ell},\beta_k)$]\label{thm:der_of_Q}
For function $u_{\ell}: \Omega\rightarrow [0,1]$, $\mathbb{X}^{(t)}_{\bm b, \beta_k},\mathbb{X}^{(t)}_{\bm d, \beta_k}$ is the set of birth and death critical points of $u_{\ell}$, we have 
\begin{align*}
    \frac{\delta \mathcal{Q}_{\varepsilon}}{\delta u_{\ell}}(x) = \int_{\mathbb{X}^{(t)}_{\bm b, \beta_k}} k_M(y,x)  \mathrm{d}y - \int_{\mathbb{X}^{(t)}_{\bm d, \beta_k}} k_m(y,x)  \mathrm{d}y,
\end{align*}
where $k_M(y,x)$ and $k_m(y,x)$ is defined at \cref{def:smooth_ker}.
\end{theorem}

\begin{figure}[!ht] 
  \centering
  \subcaptionbox*{Without width: The result of minimization energy \cref{eq:phenergy}.}
    {\includegraphics[width=0.95\linewidth]{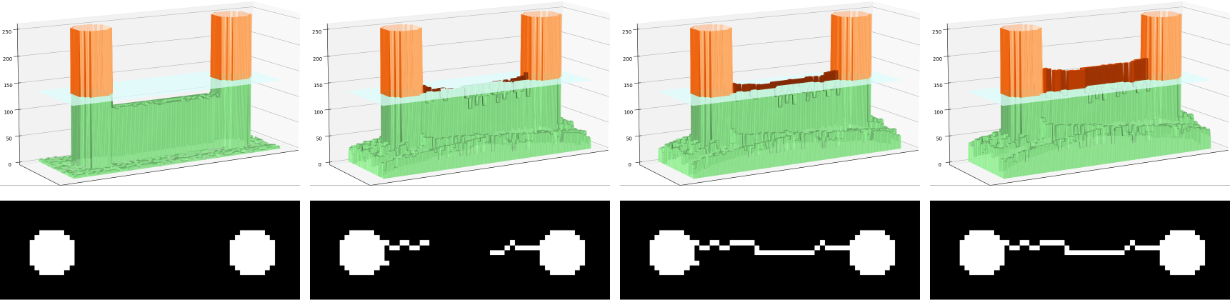}}\\
  \subcaptionbox*{Ours: The result of minimization energy \cref{eq:wtenergy}.}
    {\includegraphics[width=0.95\linewidth]{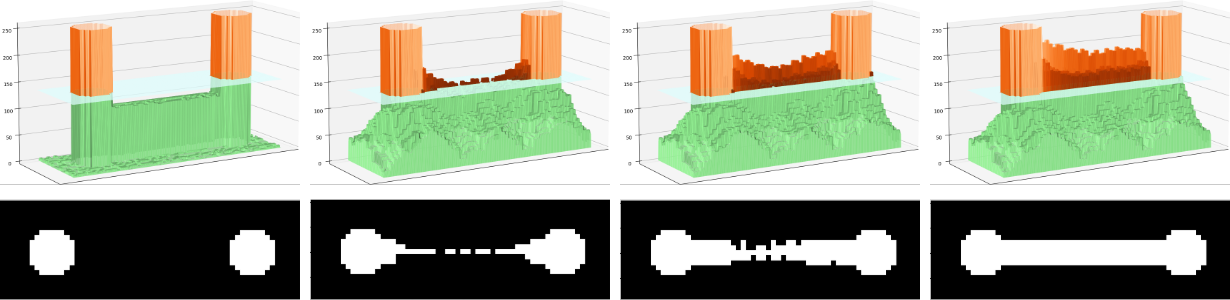}}
    \caption{Iterative process to minimize topological energy. The energy without width information \cref{eq:phenergy} has the gradient for backpropagation only at the critical point, but our width-aware topological energy \cref{eq:wtenergy} consider the neighborhood of the critical point.}
  \label{fig:idea}
\end{figure}

The proof of \cref{thm:der_of_Q} can be seen in appendix \ref{apdx:thm:der_q}. When $\varepsilon \rightarrow 0^+$, the \textcolor{black}{variation} of $\mathcal{Q}$ about $u_{\ell}$ is:
\begin{align*}
    \frac{\delta \mathcal{Q}_{\varepsilon}}{\delta u_{\ell}}(x) = \int_{\mathbb{X}^{(t)}_{\bm b, \beta_k}} k_M(y,x)  \mathrm{d}y - \int_{\mathbb{X}^{(t)}_{\bm d, \beta_k}} k_m(y,x)  \mathrm{d}y, 
\end{align*}
where
\begin{align*}
    k_{M}(y,x) = \left\{\begin{array}{cl}
1, & x = \arg\max_{z \in \mathbb{B}(y,r)} u_{\ell}(z), \\
0, & else.
\end{array}\right.\\
    k_{m}(y,x) = \left\{\begin{array}{cl}
1, & x = \arg\min_{z \in \mathbb{B}(y,r)} u_{\ell}(z), \\
0, & else.
\end{array}\right.
\end{align*}
This means that its backpropagation is only at extreme points in the neighborhood of the critical points. The gradient of the extreme value of the critical points of birth is 1 and the minimum value of the critical points of death is -1. If we do not add the morphological gradient to $\mathcal{B}(u_{\ell},\beta_k)$, for the fixed critical of $u_{\ell}^t$, it has gradients only at the critical points of birth and death, with gradients of 1 and -1, respectively. The intuitive visualization results are shown in \cref{fig:idea}.

\begin{remark}\label{rmk:moti}
    For the upper level set, we constrain the outer gradient to be minimal at the birth point and the inner gradient to be minimal at the death point, in addition to the fact that we can get a concise form, the main reason is as follows: \begin{itemize}
        \item The birth point is a maxima of its neighborhood. By considering the outer gradient in a non-smooth form, it is equivalent to the original topological energy. 
        \item The birth point is the saddle point in the neighborhood. When adding the non-smooth outer gradient at the birth point, it will change the birth point corresponding to the critical value to a maximum value of its neighborhood. According to \cref{thm:der_of_Q}, this case has gradients only at the local maxima of the critical point in backpropagation, which will eventually turn the birth point into a local maxima.
    \end{itemize}

    The reason for adding the inner gradient to the death critical point is similar.  Then considering the smooth form of the inner and outer gradients, for birth points, the gradient of the backpropagation process is larger for points with larger pixel values within their neighborhood, while the opposite is true for death points. Therefore, the pixel values within the local neighborhood of the critical point change during the iteration process, instead of backpropagation of the gradient only at the critical points, thus constraining the topological features of the fused width information.
\end{remark}

In summary, for encouraged topology structures, the neighborhood of birth points has an increasing pixel value, the neighborhood of death points has a declining value, and vise versa for penalized topology structures.

\subsection{Incorporating width-aware topological preservation to variational models and deep neural networks for image segmentation}

In order to verify the effectiveness of our proposed width-aware topological energy for image segmentation problems, we apply it to traditional variational segmentation models and data-driven models.

\subsubsection{Topology energy for varitional segmentation model}
We propose the Topology Nonlocal Soft Threshold Dynamics (Topo-NLSTD) model that maintains topology and incorporates width information as 
\begin{equation}\label{eq:nlstd}
 \min\limits_{\bm u\in \mathbb{U}}\left\{\mathcal{E}(\bm u)=\underbrace{\langle-\bm o,\bm u\rangle}_{:=\mathcal{F}(\bm u)}
 +\underbrace{\gamma\langle\bm u,\ln \bm u\rangle}_{:=\mathcal{S}(\bm u)}+\underbrace{\langle \bm u, \mathcal{N}(\bm u)\rangle}_{:=\mathcal{R}(\bm u)}+\eta \mathcal{T}_{\varepsilon}(u_{\ell})\right\},
 \end{equation}
 where $\mathbb{U}$ is defined at \cref{eq:u_condition}, $\mathcal{T}_{\varepsilon}$ is defined at \cref{eq:wtenergy} \textcolor{black}{and the parameter $\beta_k$ is predetermined to satisfy the corresponding topological structure.}
$u_\ell(x)$ is one of the channels containing explicit topological features, \textcolor{black}{which requires manual pre-specification. In cases where multiple such channels exist, each with its own well-defined topology, we can impose their corresponding topological constraints. For simplicity of exposition, we impose constraints only on the topological properties of a single channel. The multi-channel case can be handled similarly by merely adding the topological constraint terms for the new channels to \cref{eq:nlstd}.} The term $\mathcal{S}$ of \cref{eq:nlstd} is to make the solution of $\bm u$ smooth, which prevents the solution from causing binarization during the iterative solution process. $\mathcal{N}$ is a vector-valued function whose $\ell$-th component function is defined by
\begin{equation} \label{eq:N}
\begin{array}{rl}
\left[\mathcal{N}(\bm u)\right]_{\ell}=&\lambda_{\ell}\displaystyle\sum_{\ell^{'}=1}^{L}   \zeta_{\ell,\ell^{'}}\int_{\Omega}w(x,y)(1-u_{\ell^{'}}(y))\mathrm{d}y.
\end{array}
\end{equation}

Compared to local STD\cite{std}, there are several important benefits: first, weight in NLSTD can integrate nonlocal information of the image; \textcolor{black}{second}, $\bm\zeta$ is a correlation matrix that is used to measure the compatibility of classes. Taking an example, if each class is independent, then $\bm\zeta$ is an identity matrix, and it reduces to the well-known Potts model. Third, by adding the topological regularization term, the model is made to constrain the topology containing the width information during the iterative solution process.

There are many choices for the weighting function $w$. Inspired by \cite{densecrf}, we \textcolor{black}{use} a Gaussian mixture model to approximate it as 
\begin{equation}\label{eq:crf}
w(x,y)=\sum_{\ell}\omega_{\ell}^0\exp\left(-\frac{||I(x)-I(y)||^2}{\alpha_{\ell}^1}-\frac{||x-y||^2}{\alpha_{\ell}^2}\right)+\omega_{\ell}^1\exp\left(-\frac{||x-y||^2}{\alpha_{\ell}^3}\right),
\end{equation}
where $\omega_{\ell}^0,\omega_{\ell}^1,\alpha_{\ell}^1, \alpha_{\ell}^2, \alpha_{\ell}^3 \ge 0 $. $\omega_{\ell}^0,\omega_{\ell}^1$ control the weights of the unitary and binary potential functions, $\alpha_{\ell}^1$ is the standard deviation of the Gaussian color kernel. The smaller the value, the weaker the interaction between pixels with large color differences. $\alpha_{\ell}^2,\alpha_{\ell}^3$ is the standard deviation of the spatial Gaussian kernel. The larger the value, the stronger the interaction between the more distant pixels.

Since the fidelity term, the regularization term of the optimization problem \cref{eq:nlstd} all contain $\bm u$, 
it is difficult to solve it directly using gradient methods and demonstrate its stability. The operator splitting method \cite{opratersplit} is suitable for such objective functions. Combined with the Lagrangian method \cite{alm}, it decomposes a complex problem into two or more simple sub-problems by introducing new operators. Here we introduce a new variable $v_\ell$ to transform the original problem into:
\begin{equation*}
    \min\limits_{\bm u\in \mathbb{U}, v_\ell} \mathcal{F}(\bm u) +  \mathcal{S}(\bm u) + \mathcal{R}(\bm u)+\mathcal{T}_{\varepsilon}(v_{\ell}), \quad \mathrm{s.t.} \quad v_{\ell}-u_{\ell} = 0.
\end{equation*}
Using the penalty function method, the above problem can be approximated by:
\begin{equation*}
    \min\limits_{\bm u\in \mathbb{U}, v_\ell} \mathcal{F}(\bm u) +  \mathcal{S}(\bm u) + \mathcal{R}(\bm u)+\mathcal{T}_{\varepsilon}(v_{\ell}) + \eta \|v_\ell-u_\ell\|_1,
\end{equation*}
with a large penalty parameter $\eta$. 
The primary reason for choosing the $L_1$ penalty term over the $L_2$ penalty term is that the dual problem of $L_1$ is linear with respect to $\bm u$, allowing the subproblem of $\bm u$ \cref{eq:iter_ccp} to have a closed form solution.
\begin{theorem}[Duality of $L_1$]\label{thm:duall1}
    \[ \|y\|_1 = \|y\|_1^{**} = \max_{\|q\|_{\infty}\le 1}\langle q, y\rangle. \]
\end{theorem}
The proof of \cref{thm:duall1} can be seen in appendix \ref{apdx:thm_duall1}. Thus, we obtain an equivalent optimization problem:
\begin{equation*}	
    \min\limits_{\bm u\in \mathbb{U}, v_\ell} \max\limits_{\|q_{\ell}\|_{\infty}\le 1} \mathcal{F}(\bm u) +  \mathcal{S}(\bm u) + \mathcal{R}(\bm u)+\mathcal{T}_{\varepsilon}(v_{\ell}) + \eta \langle q_\ell, v_\ell-u_\ell \rangle.
\end{equation*}
Then we can solve this problem by alternating optimization method to split it into the following three subproblems:
\begin{equation*}
    \begin{cases}
        q_\ell^{(t+1)} = \arg\max\limits_{\|q_{\ell}\|_{\infty}\le 1}\{\langle q_{\ell}, v_\ell^{(t)}-u_\ell^{(t)}\rangle\},\\
	v_{\ell}^{(t+1)} = \arg\min\limits_{v_{\ell}}\{ \mathcal{T}_{\varepsilon}(v_{\ell})+\eta\langle q^{(t+1)}, v_{\ell}\rangle\},  \\
	\bm{u}^{(t+1)} = \arg\min\limits_{\bm{u}\in\mathbb{U}}\{ \mathcal{F}(\bm u) + \mathcal{S}(\bm u) + \mathcal{R}(\bm u) + \eta\langle q_\ell^{(t+1)}, v^{(t+1)}_{\ell}-u_{\ell}\rangle\}.
    \end{cases}
\end{equation*}

For the subproblem of $q_\ell$, \textcolor{black}{although this subproblem has a closed-form  solution, directly applying the closed-form solution in the iterative process leads to an unstable solution for the $u$-subproblem. For $u_\ell$ is a relaxation indicator function, as shown in \cref{eq:u_condition}, which ensures that $u_\ell$ is projected into the interior of the simplex during iteration. When we use the closed-form solution for the $q_\ell$-subproblem - specifically, the sign function of each component of $(v_\ell - u_\ell)$ - for iteration, it causes $u_\ell$ to be oscillate between different vertices of the simplex. Therefore, inspired by the primal-dual hybrid gradient method (PDHG) \cite{pdhg}, we adopt the following approach to solve approximately:}
\begin{align*}
    q_\ell^{(t+1)} \approx \mathrm{Proj}_{\|\cdot\|_{\infty}\le1}\{q_\ell^{(t)}+(v_{\ell}^{(t)}-u_{\ell}^{(t)})\}.
\end{align*}
\textcolor{black}{which is essentially a proximal algorithm for the $q_\ell$-subproblem. It can avoid such ``jumps" and maintains the continuity of the $q_\ell$-update.}

For the subproblem of $v_\ell$, we first \textcolor{black}{substitute $\mathcal{B}_{\varepsilon}(u_{\ell},\beta_k)$ with $\mathcal{Q}_{\varepsilon}(u_{\ell},u^{(t)}_{\ell},\beta_k)$ in \cref{eq:wtenergy}}:
\begin{equation}\label{eq:compute_topoenergy}
    \mathcal{T}_{\varepsilon}(v_{\ell},v^{(t)}_{\ell})=\sum_{k=0}^1\mu_k\left(2\mathcal{Q}_{\varepsilon}(v_{\ell},v^{(t)}_{\ell},\beta_k) - \mathcal{Q}_{\varepsilon}(v_{\ell},v^{(t)}_{\ell},0)\right).
\end{equation}
\textcolor{black}{Subsequently, we compute the gradient of the subproblem with respect to $v_\ell$:
\begin{equation}\label{eq:v_grad}
        g^{(t)} = \frac{\delta \mathcal{T}_{\varepsilon}(v_{\ell},v^{(t)}_{\ell})}{\delta v_{\ell}} + \eta q_\ell^{(t+1)}, 
\end{equation}
where the derivative of $\mathcal{T}_{\varepsilon}(v_{\ell},v^{(t)}_{\ell})$ with respect to $v_\ell$ follows from \cref{thm:der_of_Q}. Due to the long computation time of the PH-based topological loss, we employ the Adaptive Moment Estimation with Decoupled Weight Decay (AdamW) \cite{adamw} optimizer for its faster convergence} and write $v_\ell^{(t+1)} = \mathcal{A}(v^{(t)}_{\ell})$ to denote the iterative process of AdamW for ease of representation:
\begin{align*}
    &m^{(t+1)} = \rho_1 \cdot m^{(t)} + (1 - \rho_1) \cdot g^{(t)}, \\
    &r^{(t+1)} = \rho_2 \cdot r^{(t)} + (1 - \rho_2) \cdot g^{(t)} \odot g^{(t)}, \\
    &\hat{m}^{(t+1)} = \frac{m^{(t+1)}}{1 - \rho_1^{t+1}}, \quad \hat{r}^{(t+1)} = \frac{r^{(t+1)}}{1 - \rho_2^{t+1}}, \\
    &v_\ell^{(t+1)} = (1 - \tau \nu) v^{(t)}_{\ell} - \tau \left( \frac{\hat{m}^{(t+1)}}{\sqrt{\hat{r}^{(t+1)}} + \epsilon} \right), 
\end{align*}
where $m^{(t)}$ is the first order moment estimation (momentum) and $m^{(0)} = 0$, $r^{(t)}$ is second-order moment estimation (adaptive learning rate) and $r^{(0)} = 0$. The learning rate $\tau$, decay rates $\rho_1, \rho_2$, weight decay $\nu$, numerical stability constant $\epsilon$ are the hyper parameters. \textcolor{black}{In the experiments, we update the gradients using \cref{eq:v_grad} and feed them into the AdamW optimizer for iterative optimization.}

For the third subproblem related to $\bm u$, $\mathcal{F}$ and $\mathcal{S}$ are convex due to the twice Legendre transformation, and the spatial prior term $\mathcal{R}$ is concave under some mild mathematical conditions for $w, \zeta$. Thus, the functional in \eqref{eq:nlstd} is smooth with respect to $\bm u$, but is non-convex. The concave-convex procedure (CCCP) can provide an efficient energy descent algorithm. The following proposition of the spatial regularization term $\mathcal{R}$ will help in constructing the algorithm.

\begin{theorem}\label{thm:der_R}
If $\bm \zeta$ and $w$ are both symmetric positive semi-definite, then $\mathcal{R}$ is concave with respect to $\bm u$.  Specifically, $\mathcal{R}$ is the TD regular term when $\bm \zeta$ and $w$ are identity matrix and Gaussian function, respectively.
\end{theorem}

The proof of \cref{thm:der_R} can be seen in appendix \ref{apdx:thm_der_r}. Thus, the regularization term $\mathcal{R}$ is concave according to \cref{thm:der_R}. Therefore, we can get the following iteration scheme:
\begin{equation}\label{eq:iter_ccp}
\bm u^{(t+1)}=\arg\min\limits_{\bm u\in\mathbb{U}}\left\{\mathcal{F}(\bm u)+\mathcal{S}(\bm u) + \mathcal{R}(\bm u^{(t)})+\langle \bm p^{(t)},\bm u-\bm u^{(t)}\rangle - \eta\langle \bm{q}^{(t+1)},\bm{u}\rangle \right\},
\end{equation}
where $\bm{q}^{(t+1)} = (0,\cdots,q^{(t+1)}_\ell,\cdots,0)$ and $\ell$ is the index of the channel with explicit topological features. $\bm p^{(t)}\in \partial \mathcal{R}(\bm u^{(t)})$ is a subgradient of $\mathcal{R}$ at $\bm u^{(t)}$ which is given by \cref{thm:der_q}. The key idea of this iteration is to replace the concave functional $\mathcal{R}$ with its supporting hyperplane at $\bm u^{(t)}$ during each iteration. The following theorem gives the calculation formula of $p^{(t)}_{\ell}(x)$.

\begin{theorem}\label{thm:der_q}
$\mathcal{R}$ is differentiable, and
$$p^{(t)}_{\ell}(x)=\lambda_{\ell}\sum_{\ell^{'}=1}^L\zeta_{\ell,\ell^{'}}\int_{\Omega}w(x,y)(1-u^{(t)}_{\ell^{'}}(y))\mathrm{d}y-\sum_{\ell^{'}=1}^L\lambda_{\ell^{'}}\zeta_{\ell^{'},\ell}\int_{\Omega}w(y,x)u^{(t)}_{\ell^{'}}(y)\mathrm{d}y.$$
Specifically,
if $\bm \zeta$ and $w$ are symmetric, \emph{i.e.,} $\zeta_{\ell,\ell^{'}}=\zeta_{\ell^{'},\ell}, w(x,y)=w(y,x)$ and $\lambda_{1}=\lambda_{2}=\cdots=\lambda_{L}$, then 
$$p^{(t)}_{\ell}(x)=\lambda_{\ell}\sum_{\ell^{'}=1}^L\zeta_{\ell,\ell^{'}}\int_{\Omega}w(x,y)(1-2u^{(t)}_{\ell^{'}}(y))\mathrm{d}y.$$
\end{theorem}

\Cref{thm:der_q} can be proved in a similar way as in \cref{thm:der_R}. Then \cref{eq:iter_ccp} has the closed solution:
\begin{equation}\label{eq:u_solution}
    \bm{u}^{(t+1)} = Softmax(\frac{ \bm o - \bm{p}^{(t)} + \eta \bm{q}^{(t+1)}}{\gamma}).
\end{equation}

The proof can be seen in the appendix \ref{apdx:eq_u_solution}. Thus, our analysis leads to the algorithm in \cref{alg:toponlstd}. For illustrative purposes, we define the set of hyperparameters  $ \bm{\theta}_{AdamW}=(\nu, \tau, \rho_1,\rho_2, \epsilon)$,  $\bm{\theta}_{nlstd}=(\lambda, \gamma, \omega^0, \omega^1, \alpha^1, \alpha^2, \alpha^3)$, and $\bm{\theta}_{topo}=(\eta, \varepsilon, r, \mu_0, \mu_1, \beta_0, \beta_1)$.

\begin{algorithm}[!ht]
    \caption{Width-Aware Topological Iterative Algorithm for Image segmentation}\label{alg:toponlstd}
    \begin{algorithmic}
        \STATE \textbf{Input:} The image $\bm{I}$, the feature $\bm o$, the hyperparameter $\bm{\theta}_{nlstd}, \bm{\theta}_{AdamW}, \bm{\theta}_{topo}$.
        \STATE \textbf{Output:} Soft segmentation function $\bm u$.
        \STATE \textbf{Initialize:}
        \STATE \hspace{0.25cm} $w(x,y)=\sum_{\ell}\omega_{\ell}^0\exp\left(-\frac{||I(x)-I(y)||^2}{\alpha_{\ell}^1}-\frac{||x-y||^2}{\alpha_{\ell}^2}\right)+\omega_{\ell}^1\exp\left(-\frac{||x-y||^2}{\alpha_{\ell}^3}\right)$;
        \STATE \hspace{0.25cm} $\bm u^{(0)} \leftarrow Softmax(\bm o)$; \quad $v_{\ell}^{(0)} \leftarrow \arg\min\mathcal{T}_{\varepsilon}(u_{\ell},u^{(0)}_{\ell})$;
        \STATE \hspace{0.25cm} $q_{\ell}^{(0)} \leftarrow \min(\max(v_\ell^{(0)}-u_\ell^{(0)}, -1), 1)$;
        \STATE \hspace{0.25cm} \textbf{for} $t=0,1,2,\dots$
        \STATE \hspace{0.5cm} \textbf{1.} update dual variable $\bm q$:
        \[ \hat{q}_\ell^{(t+1)} \leftarrow q_\ell^{(t)}+(v_\ell^{(t)}-u_\ell^{(t)}), \]
        \[q_\ell^{(t+1)}\leftarrow \min(\max(\hat{q}_\ell^{(t+1)}, -1), 1),\]
        \[\bm{q}^{(t+1)}\leftarrow (0,\cdots,q_\ell^{(t+1)},\cdots,0);\]
        \STATE \hspace{0.5cm} \textbf{2.}  update variable $v_\ell$ by AdamW: 
        \[v_\ell^{(t+1)}\leftarrow \mathcal{A}(v_{\ell}^{(t)});\]
        \STATE \hspace{0.5cm} \textbf{3.} compute the solution by CCCP algorithm:
        \[ p^{(t)}_{\ell}(x) \leftarrow \lambda_{\ell}\sum_{\ell^{'}=1}^L\zeta_{\ell,\ell^{'}}\int_{\Omega}w(x,y)(1-2u^{(t)}_{\ell^{'}}(y))\mathrm{d}y, \]
        \[ \bm u^{(t+1)} \leftarrow Softmax\left(\frac{\bm o- \bm{p}^{(t)}+\eta \bm q^{(t+1)}}{\gamma}\right); \]
        \STATE \hspace{0.5cm} \textbf{4.} Convergence check.
        \STATE \hspace{0.25cm} \textbf{end}
    \end{algorithmic}
\end{algorithm}

With \cref{alg:toponlstd}, we are able to integrate topological information with traditional variational segmentation models. \textcolor{black}{Since the solution \cref{eq:u_solution} is in the form of softmax, and the softmax function is often used as the last layer of data-driven segmentation models.} Then, \cref{alg:toponlstd} can replace the last layer of the network by the unrolling method, in the sequential manner, to fuse the width-aware topological energy into data-driven segmentation models. However, methods based on persistent homology exhibit the problem of high computational complexity and computational time. Utilizing the unrolling method described above will greatly increase the inference time of the model; therefore, we use it as a regular term in the loss function to improve the accuracy of the segmentation.

\subsubsection{Topology energy for deep neural network segmentation models}
When using deep neural networks, for example Deep Convolution Neural Networks (DCNNs), 
Dice loss is a widely used loss function in image segmentation tasks, especially in medical image processing. Its core idea is to optimize the model by measuring the degree of overlap between the predicted results $\bm u:\Omega \rightarrow [0,1]^L$ and the true labels $\bm g:\Omega \rightarrow [0,1]^L$, where $L$ is the number of categories of segmentation results:
\begin{equation*}
\mathcal{L}_{dice}(\bm u,\bm g) = 1 - \frac{1}{L}\sum_{\ell=1}^{L}\frac{2\int_{\Omega}u_{\ell}(x)g_{\ell}(x)\mathrm{d}x}{\int_{\Omega}(u_{\ell}(x))^2\mathrm{d}x + \int_{\Omega}(g_{\ell}(x))^2\mathrm{d}x}.   
\end{equation*}
However, the general loss function does not constrain the topological properties of the segmentation objective. \textcolor{black}{Similar} to previous methods, we add the proposed topological energy with width information \cref{eq:wtenergy} as a regular term to the loss function:
\begin{equation}\label{eq:topoloss}
    \mathcal{L}(\bm u, \bm g) = \mathcal{L}_{dice}(\bm u, \bm g) + \alpha \mathcal{T}_{\varepsilon}(u_{\ell}),
\end{equation}
where $u_{\ell}$ is one channel of $\bm u$ and the channel has a well-defined topological character. \textcolor{black}{During the experimental process, we need to manually specify the channels with well-defined topological properties and formulate energy functions that align with their topological characteristics.} The DCNNs and the other training details for this loss function will be specified later in the numerical section. 

\begin{remark}
In the experiments presented in this work, topological information is incorporated solely through the loss function. While methods such as those proposed in \cite{Jia2020,std} and \cite{pottsmgnet} enable the integration of topological preservation directly into DCNN architectures — so that the network outputs inherently preserve topological structures—the associated training costs are prohibitively high. As a result, these approaches have not been included in the current study. We intend to explore them in some future research.
\end{remark}

\section{Numerical experiments} \label{sec:experiment}
In this section, we verify the effectiveness of the width-aware topological (WT) energy in maintaining topologies with width information. We utilize WT energy to test the images directly, after which the test is performed in the traditional segmentation model. Finally, we test the energy on the data-driven segmentation model. Comparison experiments with the most related methods are also conducted. We use the TopologyLayer \cite{ph} package to compute persistent homology since it is open source and can be modified. The calculations in traditional model are done under Python 3.9.13, PyTorch 1.8.1+cu101 with a NVIDIA V100 GPU, and in data-driven model is under Python 3.11.11, PyTorch 2.4.0 with a NVIDIA 4090 GPU. In addition, the hyper parameters of AdamW optimizer use the default settings, \emph{i.e.} $\rho_1=0.9,\rho_2=0.999,\epsilon=10^{-8}$ in the whole section. 

\subsection{Test for width-aware topological (WT) energy}
\begin{figure}[!ht]
  \centering
  \subcaptionbox{Image}
    {\includegraphics[width=0.25\linewidth]{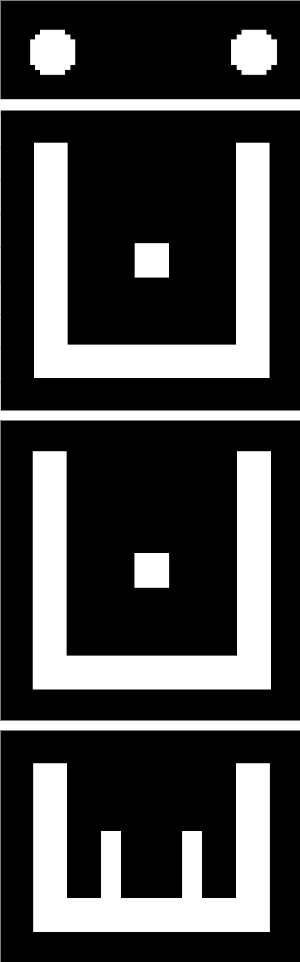}}
  \subcaptionbox{PH\cite{ph}}
    {\includegraphics[width=0.25\linewidth]{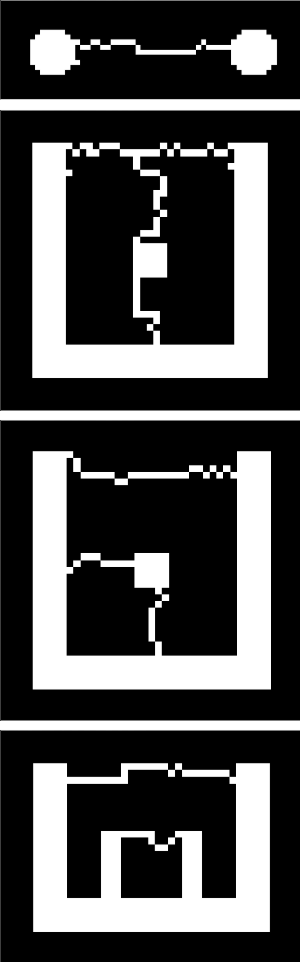}}
  \subcaptionbox{Proposed WT}
    {\includegraphics[width=0.25\linewidth]{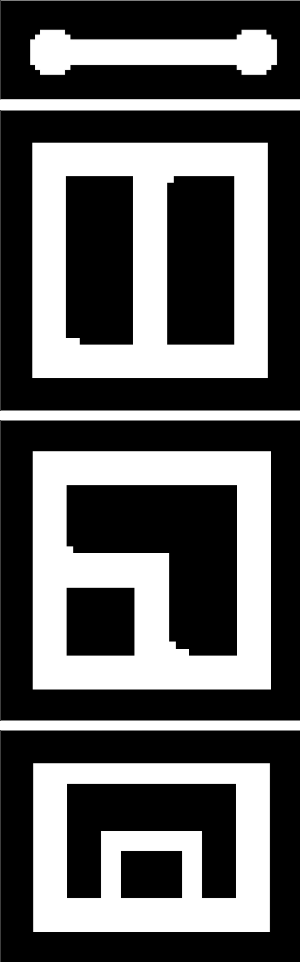}}
    \caption{Performance of topological energy. \textcolor{black}{The first row constrains one connected component and the other rows are one connect component and two holes.}}
  \label{fig:con}
\end{figure}

\begin{figure}[!ht]
  \centering
    \includegraphics[width=0.8\linewidth]{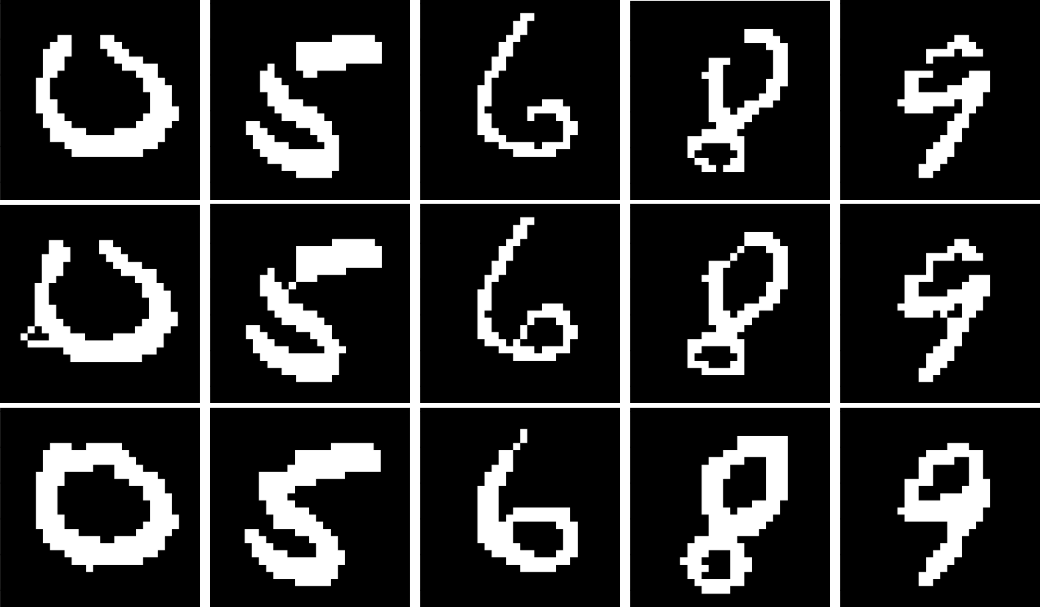}
  \caption{Performance \textcolor{black}{on} MINST dataset. The first row is the original image, second row is PH results and the thrid row is our proposed WT method.}
  \label{fig:num}
\end{figure}

\begin{figure}[!ht] 
  \centering
  \subcaptionbox*{Image}
    {\includegraphics[width=0.25\linewidth]{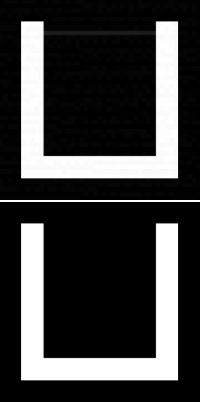}}
  \subcaptionbox*{PH\cite{ph}}
    {\includegraphics[width=0.25\linewidth]{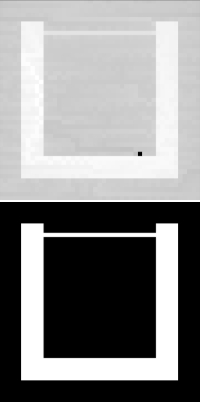}}
  \subcaptionbox*{Proposed WT}
    {\includegraphics[width=0.25\linewidth]{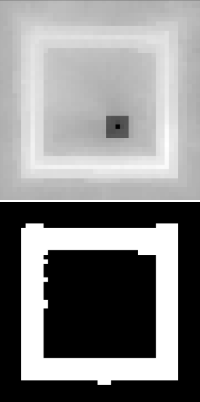}}\\
  \subcaptionbox*{Image}
    {\includegraphics[width=0.25\linewidth]{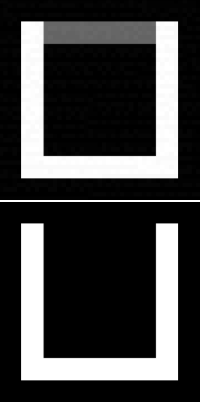}}
  \subcaptionbox*{PH\cite{ph}}
    {\includegraphics[width=0.25\linewidth]{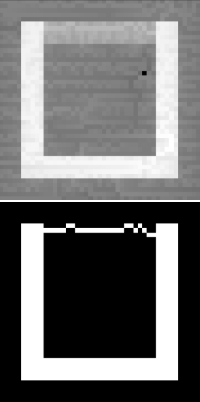}}
  \subcaptionbox*{Proposed WT}
    {\includegraphics[width=0.25\linewidth]{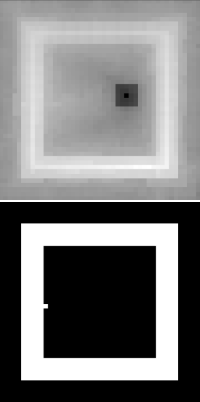}}
    \caption{Performance of topological energy to preserve one hole with different initial value. In each set of figures, the second row is the results of the first row after binarization.}
  \label{fig:hole}
\end{figure}

The first application of the WT energy is to some synthetic images and the MNIST \cite{mnist} dataset. \cref{fig:con} and \cref{fig:num} show that our method preserves the width and topology information. The results for PH are obtained by minimizing \cref{eq:phenergy} and WT is obtained by minimizing the energy \cref{eq:wtenergy} by AdamW algorithm with $\bm{\theta}_{AdamW} = (\nu, \tau) = (0.01, 0.01)$. Then, we show the results of minimizing the two energy when given different initial values, as shown in \cref{fig:hole}.

\begin{figure}
    \centering
    \includegraphics[width=1.0\linewidth]{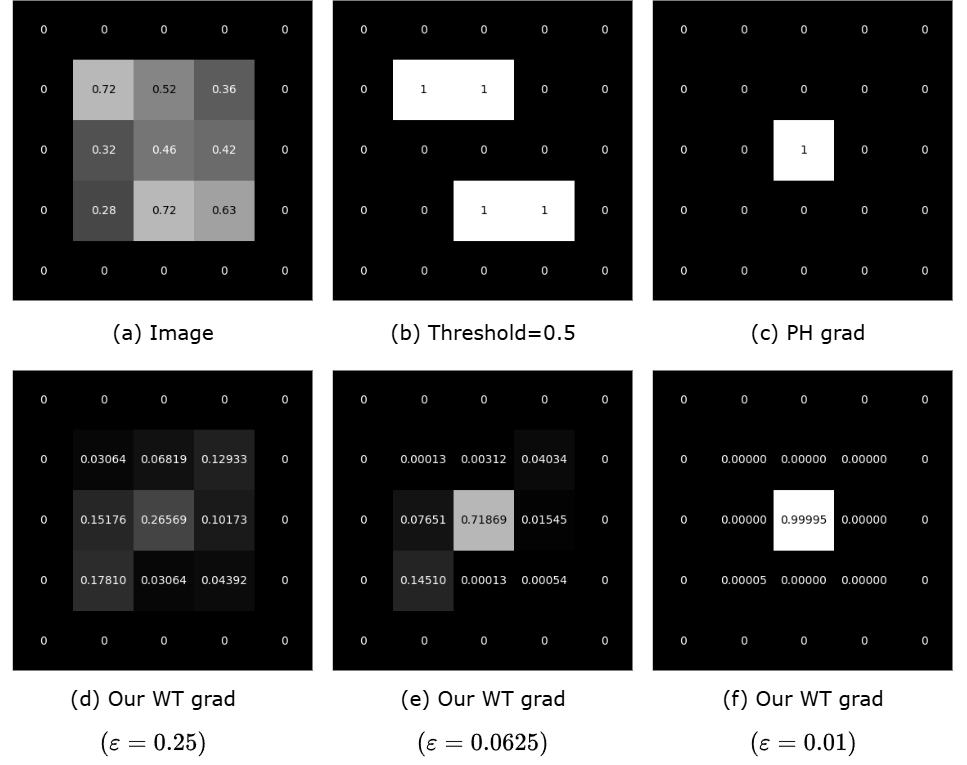}
    \caption{\textcolor{black}{Example of demonstrating the influence of parameter $\varepsilon$ on the width. (a) shows the input image, while (b) is the binarized result of (a) with a threshold of 0.5. (c)-(f) display the 
    \textbf{negative gradient visualization}. Since a simply connected image is desired, and based on the selection method of critical points, the \textbf{central pixel} corresponds to the death critical point. Here, (c) represents the negative gradient corresponding to persistent homology (PH), where the gradient exists only at the death critical point. (e)-(f) show the negative gradient of our proposed WT energy, where pixels within the structuring element neighborhood of the critical point also exhibit gradients, with values positively correlated to their pixel intensities. A larger $\varepsilon$ results in a wider width. As $\varepsilon$ gradually decreases, the gradients of pixels around the critical point progressively diminish, until the formulation degenerates into the PH energy.}}
    \label{fig:eps}
\end{figure}

\textcolor{black}{Then, based on the visualization results \cref{fig:eps}, we proceed to explain the influence of the parameter $\varepsilon$ used in the smooth morphological gradient. For the purpose of illustrating the impact of $\varepsilon$ on the width, the visualization is limited to the death cases of 0-dimensional topological structures. With a smaller $\varepsilon$, the gradient from our proposed WT energy better approximates that of persistent homology (PH), which in turn results in a narrower width.}

\subsection{Test on traditional segmentation model}
\begin{figure}[!ht] 
  \centering
  \subcaptionbox*{Synthetic image}
    {\includegraphics[width=0.2\linewidth]{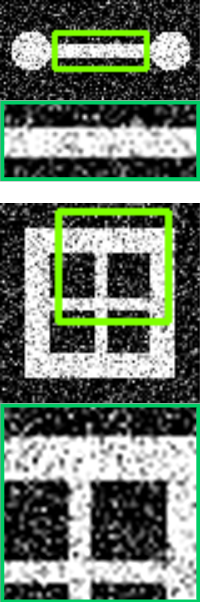}}
  \subcaptionbox*{NLSTD\cite{densecrf}}
    {\includegraphics[width=0.2\linewidth]{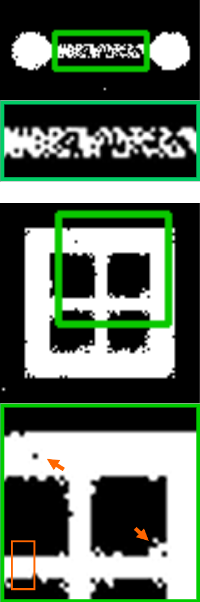}}
  \subcaptionbox*{PH\cite{ph}}
    {\includegraphics[width=0.2\linewidth]{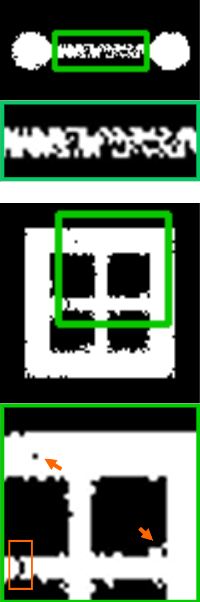}}
  \subcaptionbox*{Proposed WT}
    {\includegraphics[width=0.2\linewidth]{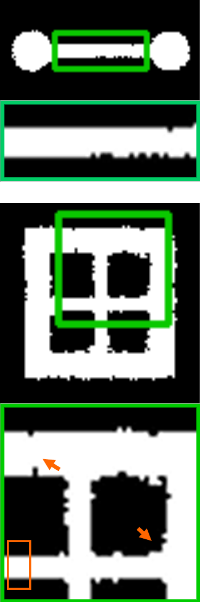}}
    \caption{Performance in synthetic image.\textcolor{black}{The upper set of diagrams is constrained to have exactly one connected component, while the lower set is constrained to be single connected and to have 4 holes.}}
  \label{fig:joystd}
\end{figure}

\begin{figure}[!ht] 
  \centering
  \subcaptionbox*{Image}
    {\includegraphics[width=0.16\linewidth]{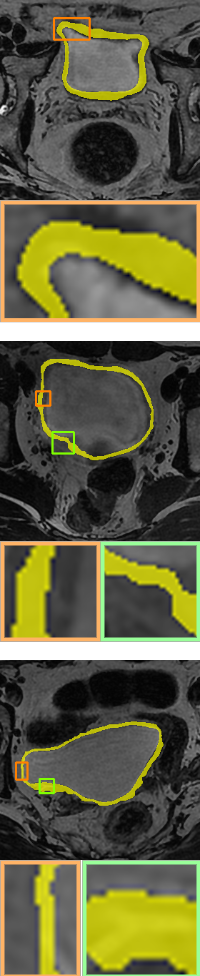}}
  \subcaptionbox*{UNet\cite{unet}}
    {\includegraphics[width=0.16\linewidth]{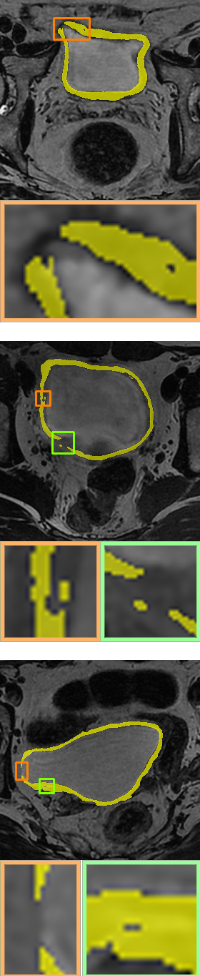}}
  \subcaptionbox*{STD\cite{std}}
    {\includegraphics[width=0.16\linewidth]{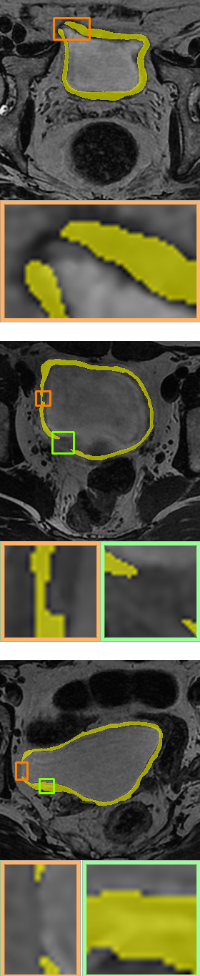}}
  \subcaptionbox*{NLSTD\cite{densecrf}}
    {\includegraphics[width=0.16\linewidth]{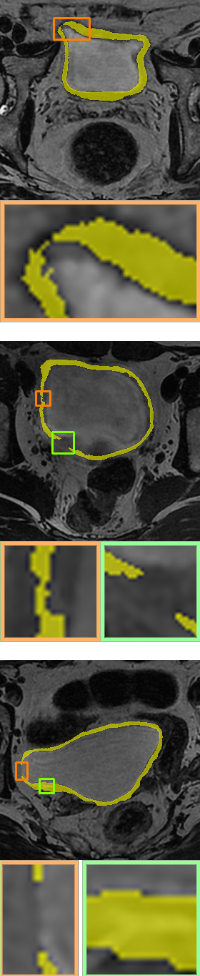}}
  \subcaptionbox*{PH\cite{ph}}
    {\includegraphics[width=0.16\linewidth]{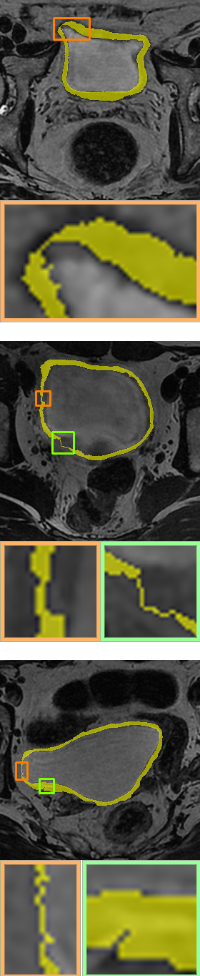}}
  \subcaptionbox*{Proposed WT}
    {\includegraphics[width=0.16\linewidth]{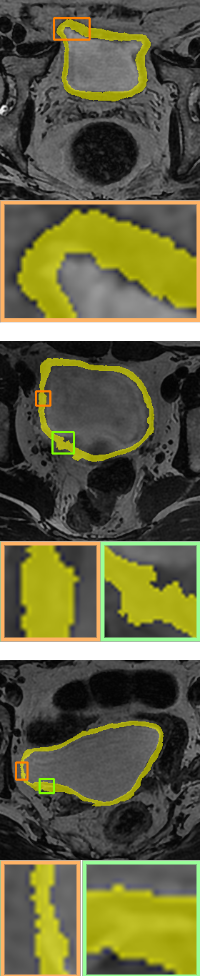}}
    \caption{Performance in ISICDM dataset. \textcolor{black}{All images in ISICDM are required to have a single connected component and one hole.}}
  \label{fig:blastd}
\end{figure}

\begin{figure}[!ht] 
  \centering
  \subcaptionbox*{Image}
    {\includegraphics[width=0.16\linewidth]{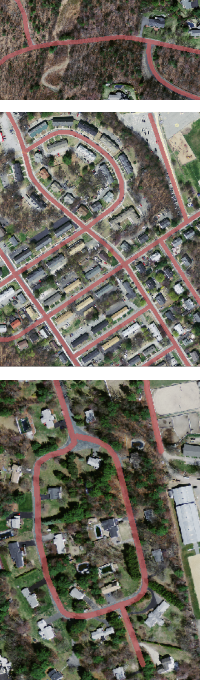}}
  \subcaptionbox*{UNet\cite{unet}}
    {\includegraphics[width=0.16\linewidth]{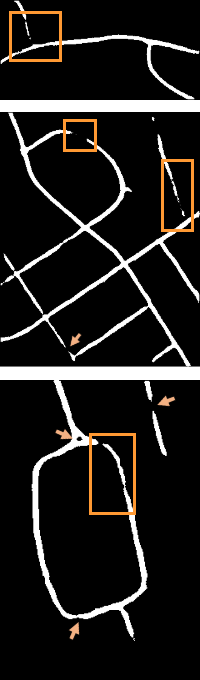}}
  \subcaptionbox*{STD\cite{std}}
    {\includegraphics[width=0.16\linewidth]{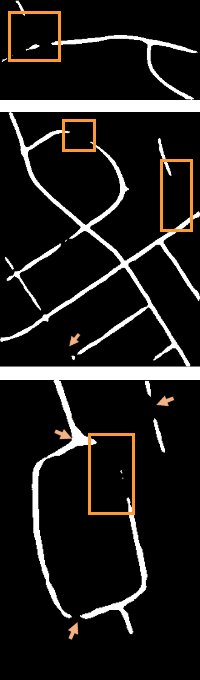}}
  \subcaptionbox*{NLSTD\cite{densecrf}}
    {\includegraphics[width=0.16\linewidth]{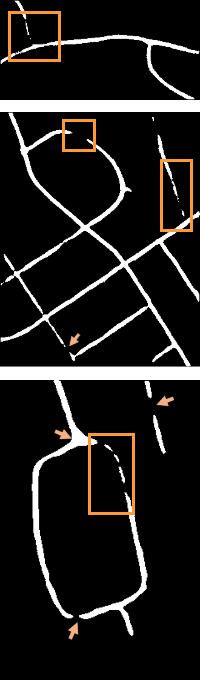}}
  \subcaptionbox*{PH\cite{ph}}
    {\includegraphics[width=0.16\linewidth]{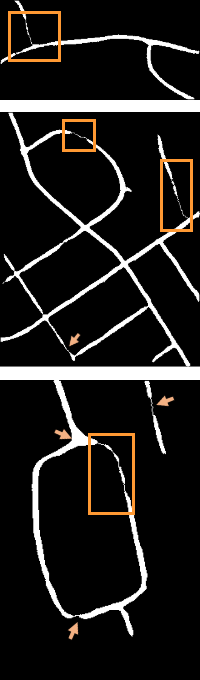}}
  \subcaptionbox*{Proposed WT}
    {\includegraphics[width=0.16\linewidth]{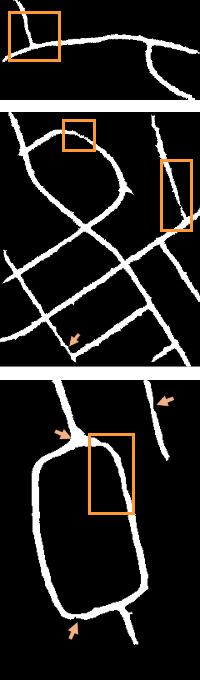}}
    \caption{Performance in Massachusetts dataset. \textcolor{black}{The first row is constrained to be single connected, the second row is constrained to have one connected component of genus three, and the third row has two connected components with one hole.}}
  \label{fig:massstd}
\end{figure}

In this section, we will add two different topological energies to Topo-NLSTD model \cref{eq:nlstd}. For convenience, when adding energy without width information \cref{eq:phenergy}, it is denoted as PH; when adding width-aware topological energy \cref{eq:wtenergy}, it is denoted as WT. \cref{fig:joystd}, \cref{fig:blastd} and \cref{fig:massstd} show the results on synthetic images, the International Symposium on Image Computing and Digital Medicine (ISICDM) 2019 dataset \cite{isicdm} and the Massachusetts Roads dataset \cite{mass}, respectively. From the results, we can see that NLSTD retains more image information compared to STD rather than performing a simple smoothing. Then, the results of PH are usually connected to a single-pixel width line or a single-pixel genus to satisfy a given topological constraint. In contrast, our proposed energy not only maintains the topological consistency well, but also fuses the width information. 

Next, we will give the experimental parameter settings in \cref{tab:paranlstd}. In \cref{fig:blastd} and \cref{fig:massstd}, the parameters of the local STD \cref{eq:std} are same as the unitary potential function in \cref{eq:crf}. In addition, $o$ in \cref{eq:nlstd} and \cref{eq:std} are the features extracted by UNet \cite{unet}.

\begin{table}[!ht] 
\centering
\caption{The hyperparameters of Topo-NLSTD}
\label{tab:paranlstd}
\begin{tabular}{lcccc}
\toprule
\multicolumn{2}{c}{Figures} & $\bm{\theta}_{nlstd}$ & $\bm{\theta}_{AdamW}$ & $\bm{\theta}_{topo}$ \\
\multicolumn{2}{c}{\quad} & $(\lambda, \gamma, \omega^0, \omega^1, \alpha^1, \alpha^2, \alpha^3)$ & $(\nu, \tau)$ & $(\eta, \varepsilon, r, \mu_0, \mu_1, \beta_0, \beta_1)$ \\
\midrule
\multirow{2}{*}{\cref{fig:joystd}} & raw 1 & (0.5, 0.3, 10, 10, 1, 3, 3) & (0.01, 0.003) & (3, 0.0625, 3, 1, 1, 1, 0) \\
  & raw 2 & (0.5, 0.3, 5, 3, 1, 3, 3) & (0.01, 0.003) & (3, 0.0625, 2, 1, 1, 1, 4) \\
\midrule
\multirow{3}{*}{\cref{fig:blastd}} & raw 1 & (0.02, 1, 5, 1, 1, 1, 1) & (0.01, 0.003) & (3, 0.0625, 2, 1, 1, 1, 1) \\
  & raw 2 & (0.02, 0.5, 10, 1, 1, 2, 2) & (0.01, 0.003) & (3, 0.0625, 2, 1, 1, 1, 1) \\
  & raw 3 & (0.02, 0.6, 10, 2, 1, 3, 3) & (0.01, 0.003) & (3, 0.0625, 2, 1, 1, 1, 1) \\
\midrule
\multirow{3}{*}{\cref{fig:massstd}} & raw 1 & (0.01, 1, 10, 0.5, 0.1, 1, 2) & (0.01, 0.003) & (5, 0.0625, 2, 1, 1, 1, 0) \\
  & raw 2 & (0.01, 1, 10, 0.5, 0.1, 1, 2) & (0.01, 0.003) & (3, 0.0625, 2, 1, 1, 1, 3) \\
  & raw 3 & (0.01, 0.1, 10, 1, 1, 3, 3) & (0.01, 0.003) & (1, 0.0625, 2, 1, 1, 2, 1) \\
\bottomrule
\end{tabular}
\end{table}

\subsection{Test on deep neural network segmentation models} 
In this section, we test the energy without width information \cref{eq:phenergy}, denoted PH, and the width-aware topological energy \cref{eq:wtenergy}, denoted WT, in the data-driven segmentation models.

\subsubsection{Datasets}
Several medical datasets are employed in our experiments. Because of the subtle differences in pixel values and the stringent requirements for geometry and detail, even small errors can lead to significant consequences. The datasets utilized for our experiments include:
\begin{itemize}
    \item Magnetic resonance imaging-based segmentation of the inner and outer bladder wall dataset \cite{isicdm} is from the Third International Symposium on Image Computing and Digital Medicine, which is called the ISICDM dataset. We select images from the dataset that are single-connected and contain one genus and divide them subsequently into training (480), validation (80) and test (120) sets. The size of each image in the dataset is 512 × 512, and the goal is to achieve accurate segmentation of the inner and outer boundaries of the bladder.
    \item The electron microscopy neuron structure segmentation dataset \cite{isbi} is provided by the challenge of neuron structure segmentation in electron microscopy (EM) stacks organized by the IEEE International Symposium on Biomedical Imaging, which is referred to as the ISBI dataset. The data consists of a set of 30 sequential slice transmission electron microscopy images of Drosophila first instar larvae ventral nerve cord (VNC) of size 512 × 512 pixels and we randomly partition the training set into training set (18), test set (6) and validation set (6) by 3:1:1.
\end{itemize}

In addition, we perform the following preprocessing on these open access datasets during the training process: 
\begin{itemize}
    \item The ISICDM dataset: Since the bladder wall is located in the middle region of the image, we center crop the entire dataset to 320 × 320 and then flip them horizontally and vertically. Finally, we normalize the images to facilitate network training.
    \item The ISBI dataset: To facilitate training and assessment of the effectiveness of topological energies, we used the cell membrane as the foreground and the cell as the background. Then, we randomly scale the images in equal proportions and flip them horizontally and vertically. Next, we apply random cropping to a size of $256 \times 256$ and finally normalize the images to facilitate network training.
\end{itemize}

\subsubsection{Evaluation indicators}
For evaluation metrics, we compare the performance of various experimental setups using three types: volumetric, boundary-based and topology-based.
\begin{itemize}
    \item Volumetric: Accuracy (Acc), Intersection Over Union (IoU) and Dice score coefficient (Dice).
    \item Boundary-based: Boundary-IoU (BDIoU) \cite{bdiou} and 95$\%$ Hausdorff distance (HD95).
    \item Topology-based: cl-Dice \cite{cldice}, 0- and 1-Betti numbers ($\beta_0$ and $\beta_1$) \cite{gudhi}.
\end{itemize}

\subsubsection{Implementation details}
 We compare the performance of the topological energy with popular encoder-decoder-based networks, such as PottsMGNet \cite{pottsmgnet}, DeepLabV3+ \cite{deeplabv3+}, SegFormer \cite{segformer} and UNet++ \cite{unet++}. The implementation of UNet++, DeepLabV3+ and SegFormer used the segmentation models PyTorch package \cite{segmodel}. In addition, the pre-training encoder weights of UNet++, Deeplabv3+ is ResNet101 and the SegFormer is MiT-B5.  We train these models with AdamW optimizer
 using the same learning strategy. The parameters of \cref{eq:topoloss} are shown as \cref{tab:paraloss}.

\begin{table}[!ht] 
\centering
\caption{The hyperparameters of Topo-NLSTD}
\label{tab:paraloss}
\begin{tabular}{cccc}
\toprule
Dataset & batch-size & $\bm{\theta}_{AdamW}:(\nu, \tau)$ & $\bm{\theta}_{topo}:(\alpha, \varepsilon, r, \mu_0, \mu_1, \beta_0, \beta_1)$ \\
\midrule
 ISICDM & 8 & (0.0001, 0.0005) & (0.00002, 0.0625, 1, 1, 1, 1, 1) \\
 ISBI & 4 &  (0.0001, 0.0005) & (0.0001, 0.0625, 1, 1, 0, 1, 0) \\
\bottomrule
\end{tabular}
\end{table}

\subsubsection{Results}

\begin{figure}[!ht] 
  \centering
  \subcaptionbox*{Ground-truth}
    {\includegraphics[width=0.19\linewidth]{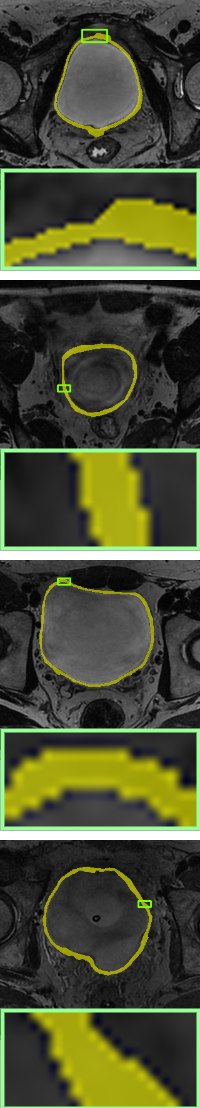}}
  \subcaptionbox*{Without-topo}
    {\includegraphics[width=0.19\linewidth]{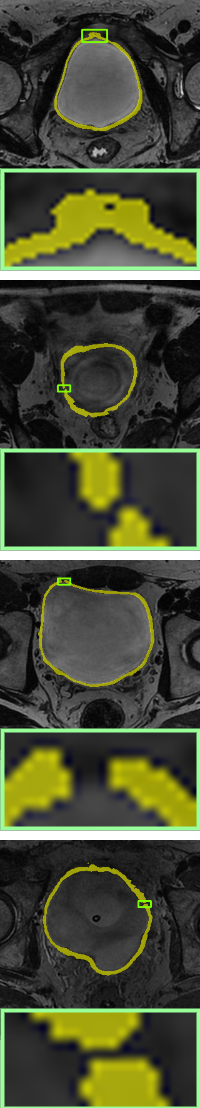}}
  \subcaptionbox*{PH\cite{ph}}
    {\includegraphics[width=0.19\linewidth]{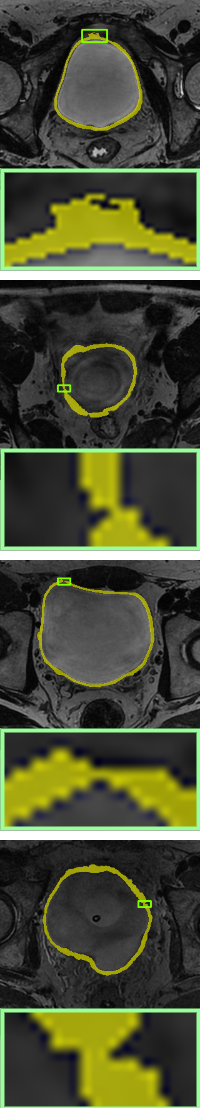}}
  \subcaptionbox*{Proposed WT}
    {\includegraphics[width=0.19\linewidth]{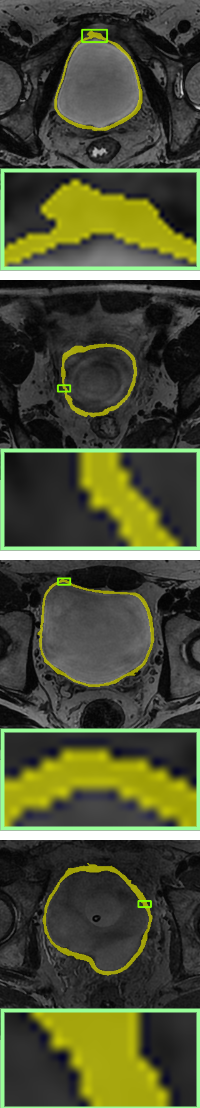}}
  \caption{Performance \textcolor{black}{on} ISICDM dataset. From top to bottom by row: PottsMGNet\cite{pottsmgnet}, DeeplabV3+\cite{deeplabv3+}, UNet++\cite{unet++}  and SegFormer\cite{segformer}.}
  \label{fig:bla-loss}
\end{figure}

\begin{figure}[!ht] 
  \centering
  \subcaptionbox*{Ground-truth}
    {\includegraphics[width=0.23\linewidth]{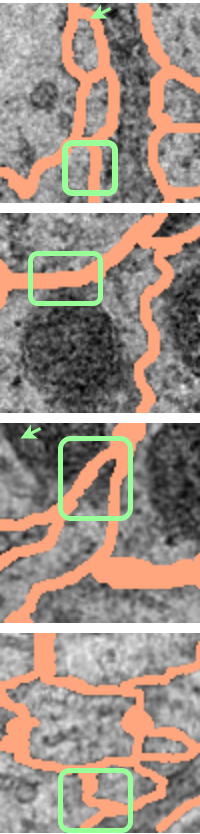}}
  \subcaptionbox*{Without-topo}
    {\includegraphics[width=0.23\linewidth]{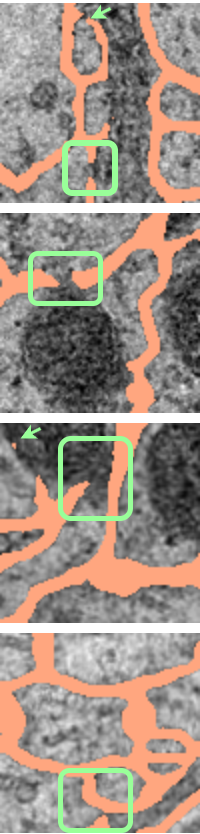}}
  \subcaptionbox*{PH\cite{ph}}
    {\includegraphics[width=0.23\linewidth]{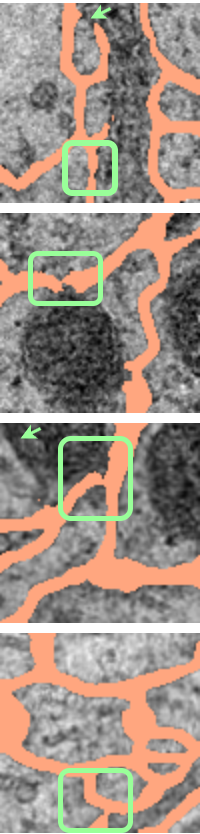}}
  \subcaptionbox*{Proposed WT}
    {\includegraphics[width=0.23\linewidth]{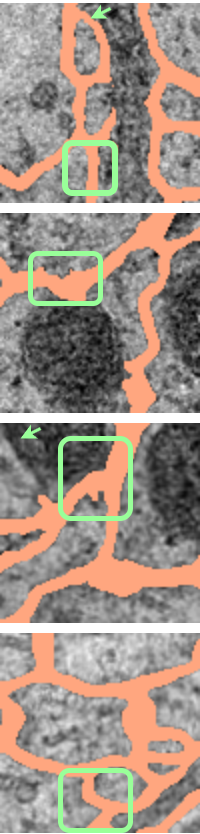}}
  \caption{Performance in ISBI dataset. From top to bottom by row: DeeplabV3+\cite{deeplabv3+}, PottsMGNet\cite{pottsmgnet}, SegFormer\cite{segformer} and UNet++\cite{unet++}.}
  \label{fig:isbi-loss}
\end{figure}

\begin{table}[!ht] 
\small
\centering
\caption{Numerical Results}
\label{tab:numresult}
\begin{tabular}{c|ccc|cc|cc}
\toprule
\multicolumn{8}{c}{ISBI dataset results} \\
\midrule
Model & Acc$\uparrow$ & Dice$\uparrow$ & IoU$\uparrow$ & BDIoU$\uparrow$ & HD95$\downarrow$ & clDice$\uparrow$ & $\beta_0$$\downarrow$ \\
\midrule
 PottsMGNet\cite{pottsmgnet} &  91.7337 & 80.8246 & 67.9062 & 67.7749 & 4.4120 & 93.1662 & 0.2930  \\
 \qquad$\sim$ + PH\cite{ph} & 91.8865 & 81.1849 & 68.4114 & 68.2860 & 4.5360 & \textbf{93.3143} & 0.3092   \\
\rowcolor{gray!16} ~~$\sim$ + WT & \textbf{92.0098} & \textbf{81.3236} & \textbf{68.6041} & \textbf{68.5000} & \textbf{4.4120} & 93.2153 & \textbf{0.2923} \\
\midrule
 DeeplabV3+\cite{deeplabv3+} & 91.4813 & 80.4784 & 67.4187 & 67.2363 & 3.9952 & 93.0107 & 0.3027 \\
 \qquad$\sim$ + PH\cite{ph} & 91.5148 & 80.8069 & 67.9082 & 67.7473 & 3.9747 & 93.3285 & 0.3001 \\
\rowcolor{gray!16} ~~$\sim$ + WT & \textbf{91.6060} & \textbf{81.0634} & \textbf{68.2531} & \textbf{68.0728} & \textbf{3.8422} & \textbf{93.6078} & \textbf{0.2910} \\
\midrule
 Unet++\cite{unet++} & 91.9790 & 81.7589 & 69.2357 & 69.0619 & 3.9573 & 93.9489 & 0.2689 \\
 \qquad$\sim$ + PH\cite{ph} & 92.0181 & 81.9812 & 69.5761 & 69.4354 & 3.9310 & 94.1658 & 0.2630 \\
\rowcolor{gray!16} ~~$\sim$ + WT &  \textbf{92.2678} & \textbf{82.0805} & \textbf{69.6741} & \textbf{69.5382} & \textbf{3.7019} & \textbf{94.3812} & \textbf{0.2520} \\
\midrule
 SegFormer\cite{segformer} & 91.6515 & 81.2004 & 68.4524 & 68.3161 & 3.9613 & 93.6786 & 0.2826 \\
 \qquad$\sim$ + PH\cite{ph} & 91.5710 & 81.1917 & 68.4583 & 68.3726 & 3.8233 & 93.8641 & 0.2865 \\
\rowcolor{gray!16} ~~$\sim$ + WT & \textbf{91.7144} & \textbf{81.2768} & \textbf{68.5622} & \textbf{68.4232} & \textbf{3.7693} & \textbf{93.9254} & \textbf{0.2813} \\
\midrule
\midrule
\multicolumn{8}{c}{ISICDM dataset results} \\
\midrule
Model & Acc$\uparrow$ & Dice$\uparrow$ & IoU$\uparrow$ & BDIoU$\uparrow$ & HD95$\downarrow$ & clDice$\uparrow$ & $\beta_1$$\downarrow$ \\
\midrule
 PottsMGNet\cite{pottsmgnet} & 99.4646 & 89.4238 & 82.0931 & 82.0962 & 3.1019 & \textbf{95.5811} & 0.4500  \\
 \qquad$\sim$ + PH\cite{ph} & 99.4665 & 89.3282 & 82.0608 & 82.0748  & 2.9676 & 95.3182 & 0.4333  \\
 \rowcolor{gray!16} ~~$\sim$ + WT & \textbf{99.4796} & \textbf{89.5576} & \textbf{82.3042} & \textbf{82.2845} & \textbf{2.7938} & 95.5554 & \textbf{0.3333} \\
\midrule
 DeeplabV3+\cite{deeplabv3+} &  99.3762 & 87.8534 & 79.3783 & 79.3316 & 2.7266 & 96.1400 & 0.4667 \\
 \qquad$\sim$ + PH\cite{ph} & 99.3733 & 87.8447 & 79.2954 & 79.2518 & 2.8483 & 96.0835 & \textbf{0.3333} \\
 \rowcolor{gray!16} ~~$\sim$ + WT & \textbf{99.4064} & \textbf{88.3391} & \textbf{80.0729} & \textbf{80.0362} & \textbf{2.5646} & \textbf{96.4410} & 0.4083 \\
\midrule
 Unet++\cite{unet++} & 99.4572 & 88.9096 & 81.3929 & 81.3363 & \textbf{2.7490} & 95.7369 & 0.1750 \\
 \qquad$\sim$ + PH\cite{ph} & 99.4583 & 88.6108 & 81.2210 & 81.2024 & 2.8620 & 95.2185 & 0.1667 \\
 \rowcolor{gray!16} ~~$\sim$ + WT &  \textbf{99.4738} & \textbf{89.0616} & \textbf{81.6956} & \textbf{81.6488} & 2.9344 & \textbf{95.7992} & \textbf{0.1250} \\
\midrule
 SegFormer\cite{segformer} &  99.4326 & 88.8135 & 80.8399 & 80.7845 & 2.4389 & 96.7139 & 0.1167 \\
 \qquad$\sim$ + PH\cite{ph} & \textbf{99.4566} & 89.2578 & 81.4288 & 81.3653 & 2.3962 & 96.9189 & 0.1833 \\
 \rowcolor{gray!16} ~~$\sim$ + WT & 99.4559 & \textbf{89.3835} & \textbf{81.6019} & \textbf{81.5443} & \textbf{2.3646} & \textbf{97.1039} & \textbf{0.0667} \\
\bottomrule
\end{tabular}
\end{table}

We compare the results of no topological constraints (without-topo), topological constraints without width (PH), and width-aware topological constraints (WT) applied to a number of data-driven models for relevant applications. As can be seen in \cref{tab:numresult}, our proposed method (rows with gray background) shows superior performance. Specifically, the improvements in volumetric demonstrate that our method has more accurate result in a pixel-level sense, while the improvements on topology-based metrics demonstrate that WT energy has an overall more precise prediction and topological similarity. For boundary-based indicators, WT performs better compared to PH in most cases due to the width information. The corresponding visualization is given in \cref{fig:bla-loss} and \cref{fig:isbi-loss}. From the visualization results, it can be seen that in data-driven image segmentation models, no topological constraints will produce obvious topological errors, even if its segmentation results have high accuracy. When constraining topological consistency using persistent homology methods, it may be possible to connect very thin lines to ensure connectivity or a consistent number of holes. We improve the PH method by smoothing the critical points in there structuring elements neighborhood, which can preserve the width information while constraining the topological properties to be consistent.

\section{Conclusions and outlooks} \label{sec:conclution}

In this paper, a novel width-aware topological energy is proposed. The core idea is to utilize the smooth morphological gradient operator to polish the critical points of persistent homology so that points undergoing topological changes no longer disappear or appear in isolation. This is because persistent homology focus only on topological information and ignore the geometrical information of the image. Then, we demonstrate the effectiveness of the proposed energy. First, by minimizing the original persistent homology topological energy and the width-aware topological energy, we visualize how points with topological changes are born or disappear, as shown in \cref{fig:idea}. Afterward, by combining with traditional segmentation models to compare the performance of energies without topological constraints, topological constraints without width information, and width-aware topological constraints, our approach performs better numerically and visually.

In future work, we consider combining the Topo-NLSTD model with data-driven models by the unrolling method, the current difficulty being that the original persistent homology requires relatively long computation times. In addition, since the current width feature is given and the width is the same for the critical point smoothing, we will later use the network to predict the width information and act on different pixels with different radii of structuring elements to achieve better results.

\appendix

\section{\textcolor{black}{An Introduction to Persistent Homology}}\label{apdx:ph}
\textcolor{black}{This section offers a review of relevant topological concepts and persistent homology, aiming to provide an intuitive understanding along with additional details. In practice, the finite field with two elements, denoted as $\mathbb{F}_2 := \mathbb{Z}/2\mathbb{Z} = \{0,1\}$, is extensively used in computational topology.}

\subsection{Homology}
Homology is a homotopy invariant of topological space, meaning that continuous deformations of space do not change the number of holes. \textcolor{black}{A $k$-dimensional simplex $[v_0, \cdots, v_k]$ is the convex combination of $k + 1$ vertices $\{v_0, v_1,\cdots, v_k \}$ and a simplicial complex $\mathbb{X}$ is a collection of simplices such that
\begin{itemize}
    \item For every simplex $\sigma$ in $\mathbb{X}$, every face $\tau \subseteq \sigma$ is also in $\mathbb{X}$.
    \item For any two simplices $\sigma_1$ and $\sigma_2$, $\tau = \sigma_1 \cap \sigma_2$ is a face of both $\sigma_1$ and $\sigma_2$.
\end{itemize}}

Then, for the $k$-dimensional complex $\mathbb{X}$, the \textcolor{black}{formula} of homology group is calculated by the chain complex $C_{k}(\mathbb{X};\mathbb{F}_2)$ and the boundary map $\partial_{k}$. \textcolor{black}{$C_{k}(\mathbb{X};\mathbb{F}_2)$ is the vector space on the field $\mathbb{F}_2$ consisting of all formal linear combinations of $k$-dimensional simplices in $\mathbb{X}$ with coefficients in $\mathbb{F}_2$. That is, each chain $c \in C_{k}(\mathbb{X};\mathbb{F}_2)$ can be written uniquely as 
$$c = \sum_{i} n_i \sigma_i,$$
where $\sigma_i$ are the $k$-simplices of $\mathbb{X}$ and $n_i \in \mathbb{F}_2 = \{0,1\}$.} The boundary map is $\partial_{k}:C_{k}\left(\mathbb{X};\mathbb{F}_2\right)\rightarrow C_{k-1}\left(\mathbb{X};\mathbb{F}_2\right)$ which \textcolor{black}{satisfies} $\partial_{k-1} \circ \partial_{k}=0$ and $\partial_{0}=0$. \textcolor{black}{Explicitly, it is defined on the basis elements (the k-simplices) for $\sigma = [v_0, \cdots, v_k]$: $$ \partial_k(\sigma) = \sum_{i=0}^{k} [v_0, \cdots, \hat{v}_i, \cdots, v_k], $$
where $\hat{v}_i$ denotes omission of the vertex $v_i$. Then, the $k$-dimension homology group is defined as a quotient group:
$$H_{k}\left(\mathbb{X}\right):=H_{k}\left(\mathbb{X};\mathbb{F}_2\right)=\frac{Z_{k}\left(\mathbb{X};\mathbb{F}_2\right)}{B_{k}\left(\mathbb{X};\mathbb{F}_2\right)}=\frac{\operatorname{Ker} \partial_{k}}{\operatorname{Img} \partial_{k+1}},$$
where $Z_{k}\left(\mathbb{X};\mathbb{F}_2\right)$, $B_{k}\left(\mathbb{X};\mathbb{F}_2\right)$} respectively represent the group of cycle and the group of boundary chain. The elements of the k-dimension homology group are equivalence classes of k-cycles modulo k-boundaries. Intuitively speaking, homology is the difference between the boundaries and cycles. \textcolor{black}{Finally, the dimension of the homology group $H_{k} (\mathbb{X})$ gives the number of $k$-dimensional topological structures of $\mathbb{X}$. Specifically, $\operatorname{dim}\left[H_{0} \left(\mathbb{X}\right)\right]$ equals the number of connected components, $\operatorname{dim}\left[H_{1} \left(\mathbb{X}\right)\right]$ encodes the number of one-dimensional holes, and higher Betti numbers capture the essential higher-dimensional features.}

\subsection{Persistent homology}

Persistent homology is the method for calculating the lifetime of topological features. In general, the topological structures with a long survival time are more important, while features that exist for a short time are considered noise. \textcolor{black}{ let $\mathbb{X}$ be a simplicial complex, and the filter function is $f : \mathbb{X} \to \mathbb{R}$. Then, $f$ satisfies the following monotonicity condition: for any simplex $\sigma \in \mathbb{X}$ and every face $\tau \subseteq \sigma$, the function must satisfy: $f(\tau) \leq f(\sigma)$. Specifically, for any scale parameter $t \in \mathbb{R}$, the set
\[
\mathbb{X}^{t} = f^{-1}[t, +\infty) = \{\sigma \in \mathbb{X} \mid f(\sigma) \geq t\},
\]
is a subcomplex of $\mathbb{X}$.} Thus, a super-level set filtration of $\mathbb{X}$ is a monotonically growing sequence induced by decreasing the value $t$: $$\emptyset=\mathbb{X}^{t_{1}} \subseteq \mathbb{X}^{t_{2}} \subseteq \cdots \subseteq \mathbb{X}^{t_{n}}=\mathbb{X}.$$ This means that the topological structure of $\mathbb{X}^{t}$ will change with the diminution of $t$. Then the addition of a $k$-dimensional simplex $\sigma$ to $\mathbb{X}^{t}$ will have two possible outcomes: 
\begin{itemize}
  \item $\partial_{k} \sigma \in \operatorname{Img} \partial_{k}:$ $ \exists~ \omega \in C_{k}\left(\mathbb{X}^{t} \backslash \sigma\right) $ , we have $ \partial_{k} \sigma=\partial_{k} \omega$. Thus, $\sigma-\omega \in \operatorname{Ker} \partial_{k} $, this makes the dimension of the group of cycle extend one, that is, \textcolor{black}{$\sigma-\omega$ generates the new homology class, and the quotient $H_{k} = \operatorname{Ker}\partial_{k} / \operatorname{Img}\partial_{k+1}$ expands by one dimension.} 
  \item $\partial_{k} \sigma \notin \operatorname{Img} \partial_{k} $: it means that the group of boundary chain will expand by one dimension. Because $\partial_{k-1} \circ \partial_{k}=0 $, we get $\partial_{k} \sigma \in \operatorname{Ker} \partial_{k-1}$. \textcolor{black}{In addition, $\partial_{k} \sigma$ is added to $\operatorname{Img} \partial_{k}$. For $ H_{k-1}=\operatorname{ker}\partial_{k-1}/{\operatorname{Img} \partial_{k}} $, $\partial_{k} \sigma$ is a generator for the removed class, and the quotient $ H_{k-1}$ will have one fewer dimension.} 
\end{itemize}

In summary, it will create or destroy topological structures when a simplex is added in filtration. Thus, one structure has birth-death time, which is denoted as a multi-set critical value pair $(b,d)$, where $b$ is the birth parameter of homology class and $d$ is the death. \textcolor{black}{Intuitively, the image can be considered as a topographic map, where the grayscale value of a pixel represents its elevation. As the water level (threshold) $t$ continuously decreases from high to low, more and more land emerges above the surface and the topology of the emerging land changes accordingly, generating a filtration of $\mathbb{X}$ - that is, a monotonically increasing sequence of sub-complexes.}

\textcolor{black}{Whenever the topology changes, a new sub-complex $\mathbb{X}^{t_i}$ is recorded, where $t_i$ is the threshold value at which the change occurs. In this framework, the birth of a 0-dimensional feature (connected component) corresponds to the emergence of a new local maximum, i.e., the creation of a new connected component, which explains why it appears first as the threshold decreases. Its death occurs when this connected component merges with another one that appeared earlier (i.e., at a higher threshold). For the birth of a 1-dimensional feature (hole): it occurs when a new cycle forms (e.g., a region of lower intensity surrounded by regions of higher intensity becomes connected as the threshold decreases to this lower value, thus forming a cycle). When this cycle is filled (e.g., when the threshold decreases to the minimum value inside the cycle, causing the hole to be filled), the hole is dead. Therefore, for 0-dimensional features, the birth point corresponds to a local maximum pixel, while the death point corresponds to a saddle point pixel. Conversely, for 1-dimensional features, the birth point corresponds to a saddle point pixel, and the death point corresponds to a local minimum pixel.}

\begin{figure}[!ht]
  \centering
    \includegraphics[width=1.0\linewidth]{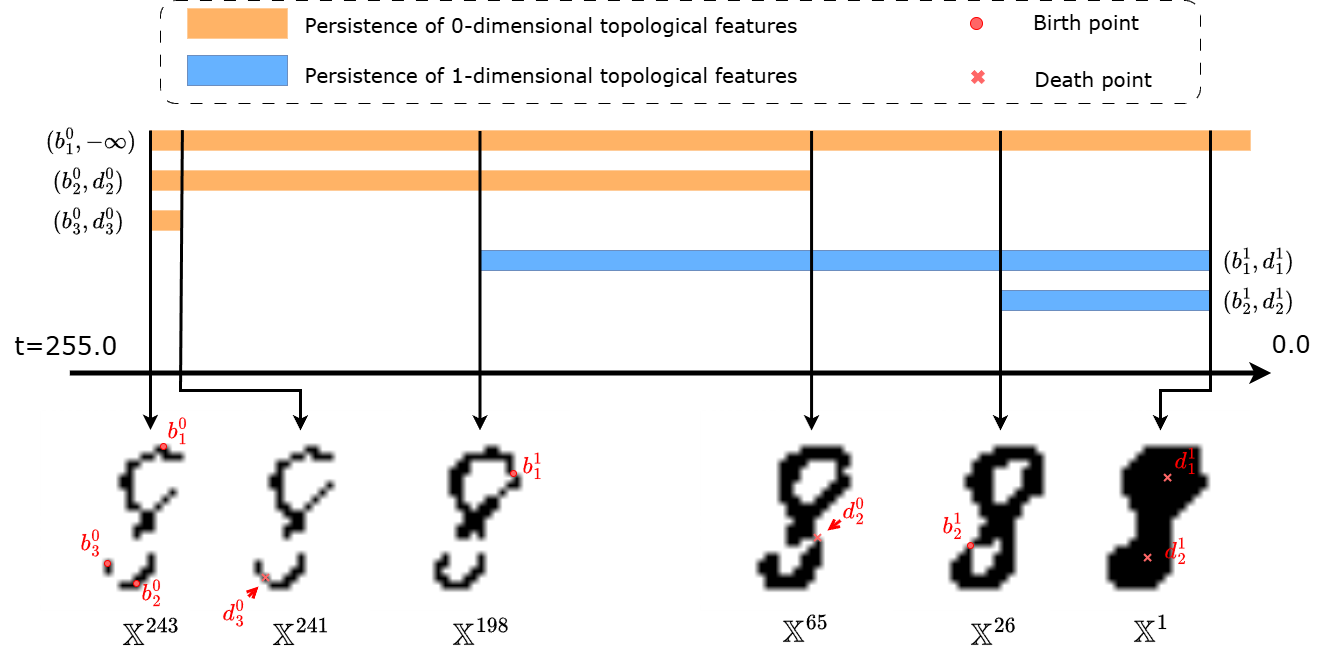}
  \caption{\textcolor{black}{A schematic diagram of persistent homology (the foreground is represented in black and the background is white). As the threshold $t$ continuously decreases, the topological structures in the image change continuously (corresponding to the black regions). Three new connected components are born at $t = 243$. When $t$ reaches 241, The two lower regions in the figure are merged, causing one of the components to die. This gives rise to a birth-death pair: $(b_3^0, d_3^0) = (243, 241)$. Then, the upper and lower regions merge into a single area at $t=65$, resulting in $(b_2^0, d_2^0) = (243, 65)$. Similarly, two distinct loops (1-dimensional features) are born at $t = 198$ and $t = 26$, respectively, and both are filled (die) at $t = 1$. Their corresponding birth-death pairs are $(b_1^1, d_1^1) = (198, 1)$ and $(b_2^1, d_2^1) = (26, 1)$.}}
  \label{fig:ph_ex}
\end{figure}

\textcolor{black}{As illustrated in \cref{fig:ph_ex}, we provide a simple example to facilitate understanding. In the experiment, similar to the simple example above, we will take image thresholds from high to low values and compute the critical points at which the topological structure of the image changes during this process. These critical values correspond to the birth and death values. Subsequently, a loss function is constructed based on the obtained birth-death pairs to enable gradient backpropagation at these critical points. The image is then updated, and the above steps are repeated. In addition, the symbols $b_j^k$ and $d_j^k$ represent the birth value and death value, respectively, of the $j$-th topological feature in dimension $k$.} Then, a $k$-dimensional persistence diagram is the union $\left\{b_{j}^{k}, d_{j}^{k}\right\}_{j=1}^{\left|\Im_{k}\right|} \cup \Delta^{\infty} $, \textcolor{black}{where $\Delta^{\infty} $ is the multi-set containing countably many copies of the diagonal $\Delta:=\left\{\left(b_{j}^{k}, b_{j}^{k}\right) \mid b_{j}^{k} \in \mathbb{R}\right\}$} and $\left|\Im_{k}\right|$ is the number of topological features. For notational convenience, the elder rule declares that if two topological structures are merged, the one with the bigger birth time will be killed. 

\section{Derivations  for \Cref{sec:smoothmorph}} \label{apdx:smoothmorph}

Remark that $\mathbb{M}=\{ \psi(x) ~|~ \psi(x) = \arg \max_{y \in \mathbb{B}(x,r)} u(y) \}$ \textcolor{black}{and $\psi(x)$ denote the coordinates of the maximum point within the neighborhood of the structuring element $\mathbb{B}(x,r)$. Next, we show the derivations for  \cref{eq:logsumexp}. Take $\widehat{x} \in \mathbb{M}$ and $u(y) \le u(\widehat{x}), ~\forall~ y \in \mathbb{B}(x,r)$:}  
\begin{align*}
    \mathcal{M}_{\varepsilon}(u) &= u(\widehat{x}) + \varepsilon \ln \int_{\mathbb{B}(x,r)} e^{\frac{u(y) - u(\widehat{x})}{\varepsilon}} \mathrm{d}y 
\le u(\widehat{x}) + \varepsilon \ln(|\mathbb{B}(x,r)|), \\
     & \Longrightarrow \quad \varlimsup\limits_{\varepsilon\rightarrow 0^{+}}\mathcal{M}_{\varepsilon}(u) \le u(\widehat{x}).
\end{align*}
According to the definition of continuity of $u$ at $\widehat{x}$, it can be proved that 
$$\varliminf\limits_{\varepsilon\rightarrow 0^{+}}\mathcal{M}_{\varepsilon}(u) \ge u(\widehat{x}) - \tau, \forall \tau>0. $$
From the arbitrariness of $\tau$, we get 
\begin{align*}
\varliminf\limits_{\varepsilon\rightarrow 0^{+}}\mathcal{M}_{\varepsilon}(u) \ge u(\widehat{x})
    \Longrightarrow \varlimsup\limits_{\varepsilon\rightarrow 0^{+}}\mathcal{M}_{\varepsilon}(u)=\varliminf\limits_{\varepsilon\rightarrow 0^{+}}\mathcal{M}_{\varepsilon}(u) &= u(\widehat{x}) = \max_{y\in \mathbb{B}(x,r)} u(y).
\end{align*}
\textcolor{black}{In the following, we show  the derivations for  \cref{eq:smooth_max}.} According to the definition of the Legendre transformation, for a fixed point $x\in\Omega$, we have
    \begin{align*}
    \mathcal{M}_{\varepsilon}^ \ast (k) := \max\limits_{u}\{ \langle k, u \rangle_\mathbb{B}(x,r) - \mathcal{M}_{\varepsilon}(u) \} 
     = \begin{cases}
    \varepsilon \langle k, \ln k\rangle_{\mathbb{B}(x,r)}, & k \in \mathbb{K}, \\
    +\infty, & else,
    \end{cases}
    \end{align*}
    where
\begin{equation*}
    \left\langle  k, \ln k \right\rangle_{\mathbb{B}(x,r)} = \int_{\mathbb{B}(x,r)} k(y) \ln k(y) \mathrm{d}y, \\
\quad
\textcolor{black}{
    \mathbb{K}=\left\{k: \mathbb{B}(x,r) \to [0,1] \mid 
    \int_{\mathbb{B}(x,r)} k(y)\mathrm{d}y = 1 \right\}.}
\end{equation*}
\textcolor{black}{
    Let $$ \mathcal{J}[u] = \int_{\mathbb{B}(x,r)} k(y) u(y) \mathrm{d}y - \varepsilon \ln \int_{\mathbb{B}(x,r)} e^{\frac{u(y)}{\varepsilon}} \mathrm{d}y. $$
    Then, 
    \begin{align*}
    \delta \mathcal{J}[u] = & \left. \frac{d \mathcal{J}[u + \tau v]}{d \tau}\right|_{\tau=0} 
 =  \int_{\mathbb{B}(x,r)} k(y) v(y) \mathrm{d}y - \varepsilon \frac{\int_{\mathbb{B}(x,r)} \frac{v(y)}{\varepsilon} e^{\frac{u(y)}{\varepsilon}} \mathrm{d}y}{\int_{\mathbb{B}(x,r)} e^{\frac{u(y)}{\varepsilon}} \mathrm{d}y} \\
    =& \int_{\mathbb{B}(x,r)} v(y) (k(y) - \frac{e^{\frac{u(y)}{\varepsilon}}}{\int_{\mathbb{B}(x,r)} e^{\frac{u(y)}{\varepsilon}}\mathrm{d}y}) \mathrm{d}y, \quad \forall  v.
    \end{align*}
    Since the original problem requires maximizing $\mathcal{J}[u]$ with respect to $u$,  the optimality condition requires
    $$ k(y) = \frac{e^{\frac{u(y)}{\varepsilon}}}{\int_{\mathbb{B}(x,r)} e^{\frac{u(y)}{\varepsilon}}\mathrm{d}y} \in [0,1], \qquad \int_{\mathbb{B}(x,r)} k(y)\mathrm{d}y = 1. $$
%
%
 Then, 
    \begin{align*}
    \mathcal{M}_{\varepsilon}^ \ast (k) 
     = \begin{cases}
    \varepsilon \langle k, \ln k\rangle_{\mathbb{B}(x,r)}, & k \in \mathbb{K}, \\
    +\infty, & else.
    \end{cases}
    \end{align*}
}
%
Following these derivations, we get
\begin{align*}
    \mathcal{M}_{\varepsilon}^{\ast\ast}(u) :&= \max\limits_{k} \{ \langle k, u \rangle_{\mathbb{B}(x,r)} - \mathcal{M}_{\varepsilon}^\ast (k)  \} 
     = \max\limits_{k\in\mathbb{K}} \{ \langle k, u \rangle_{\mathbb{B}(x,r)} - \varepsilon \langle k, \ln k\rangle_{\mathbb{B}(x,r)}  \}.
\end{align*}

\section{Proof of \cref{thm:der_of_Q}} 
\label{apdx:thm:der_q}
First, let us note 
\textcolor{black}{
    \begin{align*}
    \delta \mathcal{Q}_{\varepsilon}(u_{\ell},u^{(t)}_{\ell},\beta_k) = & \left. \frac{d \mathcal{Q}_{\varepsilon}(u_{\ell} + \tau v,u^{(t)}_{\ell},\beta_k)}{d \tau}\right|_{\tau=0} \\
    =& \int_{\mathbb{X}^{(t)}_{\bm b,\beta_k}}\int_{\Omega}  k_M(x,y) v(y) \mathrm{d}y \mathrm{d}x- \int_{\mathbb{X}^{(t)}_{\bm d,\beta_k}}\int_{\Omega} k_M(x,y) v(y) \mathrm{d}y \mathrm{d}x \\
    =& \int_{\Omega} \bm{1}_{\mathbb{X}^{(t)}_{\bm b,\beta_k}}(x) \int_{\Omega}  k_M(x,y) v(y) \mathrm{d}y \mathrm{d}x - \int_{\Omega} \bm{1}_{\mathbb{X}^{(t)}_{\bm d,\beta_k}}(x) \int_{\Omega}  k_m(x,y) v(y) \mathrm{d}y \mathrm{d}x \\
    =& \int_{\Omega} v(x) (\int_{\mathbb{X}^{(t)}_{\bm b,\beta_k}} k_M(y,x)  \mathrm{d}y -  \int_{\mathbb{X}^{(t)}_{\bm d,\beta_k}} k_m(y,x)  \mathrm{d}y )\mathrm{d}x.
\end{align*}
}
Thus,
\begin{align*}
    \frac{\delta \mathcal{Q}_{\varepsilon}}{\delta u_{\ell}}(x) =& \int_{\mathbb{X}^{(t)}_{\bm b, \beta_k}} k_M(y,x)  \mathrm{d}y - \int_{\mathbb{X}^{(t)}_{\bm d, \beta_k}} k_m(y,x)  \mathrm{d}y \\
    =& \textcolor{black}{\left\{\begin{array}{cl}\int_{\mathbb{X}^{(t)}_{\bm b, \beta_k}} \frac{e^{\frac{u_{\ell}(x)}{\varepsilon}}}{\textcolor{black}{\int_{z \in \mathbb{B}(y,r)}e^{\frac{u_{\ell}(z)}{\varepsilon}} \mathrm{d}z}} \mathrm{d}y - \int_{\mathbb{X}^{(t)}_{\bm d, \beta_k}}\frac{e^{\frac{-u_{\ell}(x)}{\varepsilon}}}{\textcolor{black}{\int_{z \in \mathbb{B}(y,r)}e^{\frac{-u_{\ell}(z)}{\varepsilon}} \mathrm{d}z}} \mathrm{d}y, & x \in \mathbb{B}(y,r), \\
    0, & x \in \Omega \setminus \mathbb{B}(y,r).
    \end{array}\right.}
\end{align*}

\section{Proof of \cref{thm:duall1}} \label{apdx:thm_duall1}
    We remark that $h(y) = \|y\|_1$, and its Legendre transform is
    \begin{equation*}
    \begin{split}
        h^{\ast}(q) &= \max\limits_{y} \{ \langle q, y \rangle - h(y) \},
         = \begin{cases}
            +\infty, \quad & \|q\|_{\infty}>1, \\
            0, \quad & \|q\|_{\infty}\le 1.
        \end{cases}
    \end{split}
\end{equation*}
    
    Then, we have the dual norm of $L_1$:
    \begin{align*}
    \begin{split}
        h^{\ast\ast}(y) &= \max\limits_{q}\{\langle y,q \rangle - h^{\ast}(q) \} = \max\limits_{\|q\|_{\infty}\le 1} \langle y,q \rangle.
    \end{split}
    \end{align*}

\section{Proof of \cref{thm:der_R}} \label{apdx:thm_der_r}

Since $\mathcal{R}$  is smooth and thus  $\partial \mathcal{R}(\boldsymbol{u})=\{\delta \mathcal{R}(\boldsymbol{u})\}$ .  Let us first calculate the directional derivative:

\begin{align*}
  \left.\frac{\mathrm{d} \mathcal{R}(\boldsymbol{u}+\tau \boldsymbol{v})}{\mathrm{d} \tau}\right|_{\tau=0} =& {\langle \bm v, \mathcal{N}(\bm u)\rangle} + {\langle \bm u, \frac{\mathrm{d}\mathcal{N}(\bm u+\tau \bm v)}{\mathrm{d}\tau}\rangle}\\
  =& {\langle \bm v, \mathcal{N}(\bm u)\rangle} + \sum_{\ell=1}^{L}\displaystyle \int_{\Omega} u_{\ell}(x) \sum_{\ell^{'}=1}^{L}  \lambda_{\ell}~ \zeta_{\ell,\ell^{'}}(\int_{\Omega}w(x,y)(-v_{\ell^{'}}(y))\mathrm{d}y)\mathrm{d}x\\
  =& {\langle \bm v, \mathcal{N}(\bm u)\rangle} - \displaystyle \sum_{\ell=1}^{L} \int_{\Omega} v_{\ell}(y) \sum_{\ell^{'}=1}^{L} \lambda_{\ell^{'}} ~ \zeta_{\ell^{'},\ell} (\int_{\Omega}w(y,x)(u_{\ell^{'}}(x))\mathrm{d}x)\mathrm{d}y\\
  =& {\langle \bm v, \mathcal{N}(\bm u) - \sum_{\ell^{'}=1}^{L} \lambda_{\ell^{'}}~ \zeta_{\ell^{'},\ell} \int_{\Omega}w(y,x)u_{\ell^{'}}(x)\mathrm{d}x \rangle}.
\end{align*}
According to the variational equation $\left.\frac{\mathrm{d} \mathcal{R}(\boldsymbol{u}+\tau \boldsymbol{v})}{\mathrm{d} \tau}\right|_{\tau=0}=\langle\boldsymbol{v}, \delta \mathcal{R}(\boldsymbol{u})\rangle$, we have $\delta \mathcal{R}(\boldsymbol{u})=\mathcal{N}(\bm u) - \sum_{\ell^{'}=1}^{L} \lambda_{\ell^{'}}~ \zeta_{\ell^{'},\ell} \int_{\Omega}w(y,x)u_{\ell^{'}}(x)\mathrm{d}x$ and $\mathcal{N}(\bm u)$ is defined at \cref{eq:N}. Besides, because $\bm \zeta$ and $w$ are symmetric semi-positive definite matrices, $\delta^{2} \mathcal{R}(\bm u) \le 0$.

\section{Proof of \cref{eq:u_solution}} \label{apdx:eq_u_solution}
We assume  $\bm u$ satisfing \cref{eq:u_condition}.  For the solution of  \cref{eq:iter_ccp} , we use  the Lagrange multiplier method. The corresponding Lagrangian multiplier and optimality condition are:
    \begin{align*}
        \mathcal{L}(\bm{u}, \widehat{q}) =& \sum_{i = 1}^{L} \int_{\Omega}\left(-o_{i}(x)+ \gamma \ln u_{i}(x) + p^{(t)}_i(x) - \eta q^{(t+1)}_i(x) \right) u_{i}(x) \mathrm{d} x \\
        &+ \int_{\Omega} \widehat{q}(x)\left(1-\sum_{i = 1}^{L} u_{i}(x)\right) \mathrm{d} x ,\\
        \frac{\partial \mathcal{L}}{\partial u_{i}} =& -o_{i} -\hat{q}+ \gamma\ln u_{i}+ \gamma + p_i^{(t+1)} - \eta q_i^{(t)} = 0  \Longrightarrow  u_i = e^{\frac{o_i + \hat{q} - p^{(t)}_i + \eta q_i^{(t+1)}}{\gamma} -1}.
    \end{align*}
    From the optimality condition, we get 
    \begin{align*}
         &1 = \sum_{i = 1}^{L} u_{i}(x) = e^{\frac{\hat{q}}{\gamma}} (\sum_{i = 1}^{L} e^{\frac{o_i - p^{(t)}_i + \eta q_i^{(t+1)}}{\gamma} -1}). \\
        \Longrightarrow \quad &e^{\frac{\hat{q}}{\gamma}} = \frac{1}{\sum_{i = 1}^{L} e^{\frac{o_i - p^{(t)}_i + \eta q_i^{(t+1)}}{\gamma} -1}}. \\
        \Longrightarrow \quad &\bm{u}^{(t+1)} = Softmax(\frac{ \bm o - \bm{p}^{(t)} + \eta \bm{q}^{(t+1)}}{\gamma}) .
    \end{align*}

\bibliographystyle{siamplain}
\bibliography{references_doi}

\end{document}

%% file: ex_shared.tex
\usepackage{lipsum}
\usepackage{amsfonts}
\usepackage{graphicx}
\usepackage{epstopdf}
\usepackage{algorithmic}
\usepackage{bm}
\usepackage{subcaption}
\usepackage[table]{xcolor} 
\usepackage{threeparttable} 
\usepackage{booktabs}
\usepackage{multirow}
\ifpdf
  \DeclareGraphicsExtensions{.eps,.pdf,.png,.jpg}
\else
  \DeclareGraphicsExtensions{.eps}
\fi

\usepackage{enumitem}
\setlist[enumerate]{leftmargin=.5in}
\setlist[itemize]{leftmargin=.5in}

\newsiamremark{remark}{Remark}
\newsiamremark{hypothesis}{Hypothesis}
\crefname{hypothesis}{Hypothesis}{Hypotheses}
\newsiamthm{claim}{Claim}

\title{
Topology-Guaranteed Image Segmentation: Enforcing Connectivity, Genus, and Width Constraints
\thanks{
\funding{This work was funded by
the National Natural Science Foundation of China (No. 42293272, 12371527)
and the Beijing Natural Science Foundation (No. 1232011).}}
}

\author{Wenxiao Li\thanks{Laboratory of Mathematics and Complex Systems (Ministry of Education of China), School of Mathematical
Sciences, Beijing Normal University, Beijing, 100875, China. 
  (\email{wxli@mail.bnu.edu.cn}).}
\and Xue-Cheng Tai\thanks{NORCE Norwegian Research Centre, Nygardstangen, NO-5838 Bergen, Norway.
  (\email{xtai@norceresearch.no}).}
\and Jun Liu\footnotemark[2] \thanks{The corresponding author. (\email{jliu@bnu.edu.cn}).}}

\usepackage{amsopn}
